\pdfoutput=1
\documentclass[10pt,letterpaper]{article}
\usepackage[letterpaper,left=0.8125in,right=0.8125in,top=1in,bottom=1.125in]{geometry}
\usepackage{times,graphicx,amssymb,booktabs,float,placeins,trimclip}
\usepackage[dvipsnames]{xcolor}
\usepackage[hyphens]{url}
\usepackage[numbers,sort&compress]{natbib}
\usepackage[format=plain,labelformat=simple,labelsep=period,font=small]{caption}
\usepackage[shortlabels,inline]{enumitem}
\setlist[itemize]{noitemsep,leftmargin=*,topsep=0em}
\setlist[enumerate]{noitemsep,leftmargin=*,topsep=0em}
\usepackage{titlesec}
\titleformat{\section}{\large\bfseries}{\thesection.}{0.4em}{}
\titlespacing*{\section}{0pt}{10pt plus 2pt minus 2pt}{7pt}
\titleformat{\subsection}{\fontsize{11}{13}\selectfont\bfseries}{\thesubsection.}{0.4em}{}
\titlespacing*{\subsection}{0pt}{8pt plus 2pt minus 2pt}{5pt}

\usepackage{amsmath,amsfonts,bm}

\def\eqref#1{equation~\ref{#1}}

\def\1{\bm{1}}

\DeclareMathAlphabet{\mathsfit}{\encodingdefault}{\sfdefault}{m}{sl}
\SetMathAlphabet{\mathsfit}{bold}{\encodingdefault}{\sfdefault}{bx}{n}

\usepackage[breaklinks,hidelinks]{hyperref}
\hypersetup{hypertexnames=false,pdftitle={Phase-Aware Video Generation for Physics-Grounded Dynamics and Interactions},pdfauthor={Jingfeng Ou, Kun Wang, Rui Zhao, Jingwei Guan, Limin Wang, Chao Dong, Xingyu Zeng},pdfsubject={Preprint}}

\newcommand{\papertitle}{Phase-Aware Video Generation for Physics-Grounded Dynamics and Interactions}
\title{\papertitle}
\author{Jingfeng Ou\textsuperscript{a,b},
Kun Wang\textsuperscript{c},
Rui Zhao\textsuperscript{c},
Jingwei Guan\textsuperscript{d}, \\[1pt]
Limin Wang\textsuperscript{a},
Chao Dong\textsuperscript{b},
Xingyu Zeng\textsuperscript{b,*}}
\newcommand{\authoraffiliations}{\textsuperscript{a}School of Computer Science, Nanjing University.\\[-1pt]
    \textsuperscript{b}Artificial Intelligence Research Institute, Shenzhen University of Advanced Technology.\\[-1pt]
    \textsuperscript{c}SenseTime.\\[-1pt]
    \textsuperscript{d}Shenzhen Technology University.\\[-1pt]
    \textsuperscript{*}Corresponding author: Xingyu Zeng.}
\date{}
\makeatletter
\renewcommand{\@maketitle}{%
  \begin{center}
    {\LARGE\bfseries \@title\par}
    \vspace{11pt}
    {\normalsize \@author\par}
    \vspace{5pt}
    {\small \authoraffiliations\par}
    \vspace{4pt}
  \end{center}%
}
\makeatother

\newcommand{\pavgroot}{.}
\newif\ifshownewrev
\shownewrevfalse
\definecolor{newrevblue}{RGB}{35,110,180}
\newcommand{\revnew}[1]{#1}
\newcommand{\revold}[1]{}
\newcommand{\revnum}[1]{#1}
\newcommand{\ojf}[1]{#1}

\newcommand{\wk}[1]{}
\begin{document}
\maketitle
\raggedbottom
\begin{abstract}
Generating physically plausible videos for solid--gas dynamics is challenging as different phases exhibit distinct dynamics yet remain coupled through physical interactions. We present \textbf{PAVG}, a \emph{Phase-Aware Video Generator} for solid--gas dynamics and interactions. It employs a dual-branch architecture to explicitly model the distinct dynamics of solids and gases, while spatiotemporal cross-attention captures their physical interactions. This design enables PAVG to preserve phase-specific motion characteristics while producing physically consistent responses across phases. To facilitate this task, we further construct a simulation corpus comprising over 700K physical trajectories across diverse solid, gas, and solid--gas interaction scenarios. Extensive evaluations demonstrate that our PAVG produces videos with improved motion adherence, physical plausibility, and visual quality compared with existing approaches.
\end{abstract}

\begin{figure*}[t]
\centering
\includegraphics[width=\textwidth]{\pavgroot/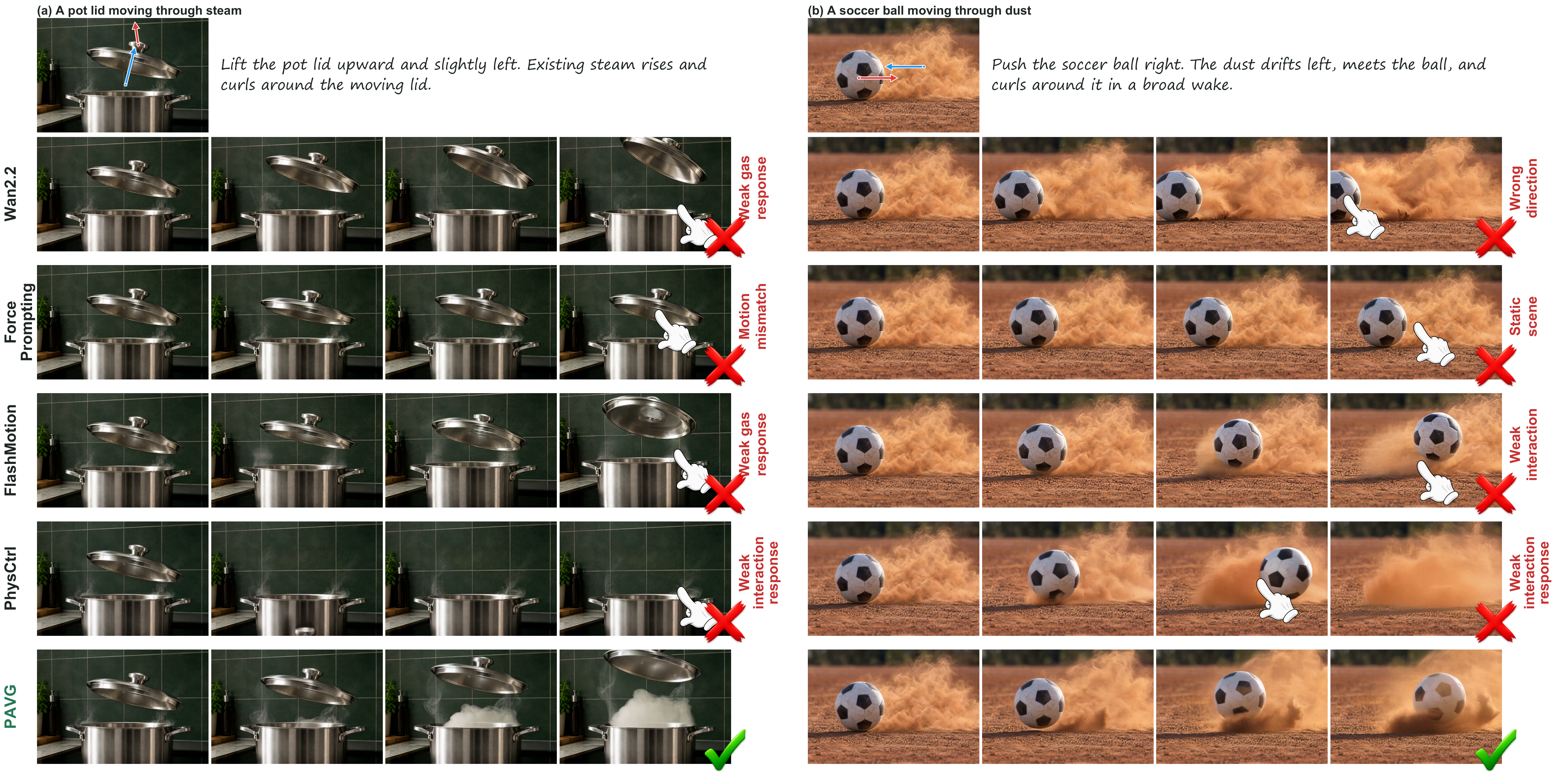}
\caption{\revnew{\textbf{Solid--gas video comparisons.}
PAVG captures steam motion around a lifted lid and dust responses to a moving soccer ball.
Marked baseline errors include weak interaction responses from PhysCtrl.
Red and blue arrows indicate solid forces and initial gas velocities.}}
\label{fig:video-teaser}
\end{figure*}

\section{Introduction}
\label{sec:intro}


Many real-world events involve multiple physical phases \footnote{In this paper, \emph{phase} denotes a modeled state of matter, such as solid or gas, rather than a phase transition.} and their interactions.
\revold{As illustrated in Figure~\ref{fig:video-teaser}, a tennis ball moving right through existing dust can deflect the leftward-drifting plume and induce local curling around the ball.}\revnew{As illustrated in Figure~\ref{fig:video-teaser}, lifting a pot lid changes the surrounding steam flow, while a soccer ball moving right through leftward-drifting dust deflects the plume around it.}
Given an initial image and an external physical intervention, we aim to generate the subsequent video with both high visual fidelity and physically plausible dynamics.
This setting requires modeling not only how each phase evolves, but also how motion in one phase influences another, such as when a moving solid induces or alters the motion of surrounding gas.

Large-scale video generation models have achieved remarkable visual fidelity and increasingly sophisticated motion generation ~\citep{sora,blattmann2023stable,xing2024dynamicrafter,yang2024cogvideox, wu2025hunyuanvideo15,wan2025wan22}.
However, visual plausibility does not guarantee physical plausibility, and recent evaluations reveal persistent failures in capturing physically plausible motion and interactions ~\citep{bansal2024videophy,bansal2025videophy}.
Such failures become particularly evident in multi-phase scenes: \ojf{\revold{in Figure~\ref{fig:video-teaser}, the baselines move the ball in the wrong direction, leave the dust stationary, or move both phases together without a visible interaction.}\revnew{in Figure~\ref{fig:video-teaser}, the baselines show weak steam responses to the moving lid, lid motion inconsistent with the prompt, incorrect ball direction, an almost static scene, or insufficient ball--dust interaction.}}
Thus, independently plausible phase dynamics are insufficient; the generated motions must also be consistent with their interactions.

Existing physics-grounded video generation approaches address this limitation by incorporating physical dynamics through either explicit simulation or learned physical priors~\citep{xie2024physgaussian,jiang2024vr,liu2024physgen, chen2025physgen3d,tan2024physmotion,physctrl,su2026physicsgroundedfluid,zhang2024physdreamer}.
The simulation-based approaches, however, are computationally expensive and sensitive to numerical settings, such as time step and spatial resolution.
They always require repeated solves and parameter adjustment for controllable generation~\citep{jiang2016material,macklin2016xpbd, modi2024simplicits,liu2024physgen,tan2024physmotion}.
In contrast, learned physical priors provide a more scalable alternative~\ojf{\citep{physctrl}}, but multi-phase scenes introduce a fundamental modeling challenge: different phases exhibit distinct dynamics, while their evolution is coupled through physical interactions~\ojf{\citep{zhang2026hgatsolver,tao2026fisale}}.
Solids may exhibit rigid or deformable motion, whereas gases involve velocity transport, dissipation, pressure, and buoyancy ~\citep{stomakhin2013material,macklin2016xpbd, modi2024simplicits,mantaflow,ummenhofer2020lagrangian,winchenbach2024sfbc, baumgarten2021coupled}.
A suitable model therefore needs to preserve phase-specific dynamics while explicitly capturing cross-phase interactions.

\ojf{We present \textbf{PAVG} (\emph{Phase-Aware Video Generation}), a framework for physics-grounded video generation of single-phase motion and solid--gas interactions.}
Given an initial image and an external intervention, PAVG lifts the image to phase-specific 3D point representations and predicts their future trajectories, which are then injected into a pretrained video generator as motion controls.
Instead of performing online physical simulation, PAVG learns physical trajectory priors directly from data.
Its dual-branch architecture separately models solid and gas dynamics, while spatiotemporal cross-attention couples the branches to capture their interactions.
We further employ a two-stage training strategy, first learning single-phase motion priors and then cross-phase interactions.
To support this setting, we construct a dataset covering diverse solid, gas, and solid–gas interaction scenarios.
\ojf{Experiments demonstrate that PAVG achieves better motion adherence, greater physical plausibility, and higher video quality than other approaches.}

Our contributions are:
\begin{itemize}
\item \ojf{We propose \textbf{PAVG} (\emph{Phase-Aware Video Generation}), a framework for physics-grounded dynamics and interactions that models single-phase motion and solid--gas interaction through a shared point-trajectory interface for physically controllable video generation.}
\item We introduce a dual-branch trajectory predictor with spatiotemporal cross-attention to capture solid-gas interactions while supporting both single- and cross-phase motion prediction.
\item We construct a trajectory dataset and a held-out benchmark covering diverse single-phase motions and solid--gas interactions.
\item We systematically evaluate trajectory prediction and video generation across single-phase and cross-phase settings, demonstrating \ojf{better adherence to motion instructions, greater physical plausibility, and higher video quality}.
\end{itemize}

\section{Related Work}
\label{sec:related-work}

\subsection{Controllable Video Generation}

Controllable video generation augments general-purpose video models~\citep{wu2025hunyuanvideo15,wan2025wan22,yang2024cogvideox} with object, region, or multi-entity motion controls~\citep{yin2023dragnuwa,wang2024motionctrl,geng2024motion,zhang2024tora,fu20243dtrajmaster,wu2024draganything,wang2024objctrl,zhang2025motionpro,li2025anyi2v,li2025magicmotion}, including point trajectories~\citep{ati2025,flextraj2025,wanmove2025,li2026flashmotion}, 3D-aware conditions~\citep{wang2025levitor,chen2025perceptioncontrol,gu2025das,zheng2026versecrafter,zhang2026symphomotion}, and structured guidance~\citep{choo2026qwerty,chen2026stance,chen2026physstream}. Recent work also models passive or rigid-body responses~\citep{liu2026diver,romero2026kinemask}. PAVG instead predicts phase-specific trajectories from material properties and applied forces before synthesis.

\subsection{Physics-Grounded Video Generation}

Physics-grounded video generation incorporates physical structure through simulation or learned priors~\citep{yang2026visualworldsurvey}. Simulation-based methods combine solvers with neural scenes or generative models~\citep{xie2024physgaussian,jiang2024vr,liu2024physgen,tan2024physmotion,chen2025physgen3d,li2025wonderplay,zhao2025physsplat,psivg2026,zhang2026physchoreo,su2026physicsgroundedfluid}; other methods learn or expose physical controls~\citep{spitznagel2025physicsgen,physctrl,forceprompting2025,goalforce2026,streamforce2026,yuan2026newtongen,narayanan2026phyco,shen2026phantom,zhang2024physdreamer} or guide generation through physical reasoning and alignment~\citep{yang2025vlipp,xue2025phyt2v,pathak2026physvid,li2026deform,li2026physalign_eccv,wang2026sift,yang2026physpo,cai2026phygdpo,zhang2026physrvg,cheng2026physrag,nguyen2026physplan,feng2026newton}. PAVG uses a shared trajectory interface to preserve distinct solid and gas dynamics and model their interaction before synthesis.

\begin{figure*}[t]
\centering
\input{\pavgroot/figures/overview_notation.tex}
\caption{\revnew{PAVG pipeline: (a) phase-specific point-cloud initialization; (b) trajectory prediction with CPI; (c) trajectory-guided video generation.}}
\label{fig:overview}
\end{figure*}

\subsection{Dynamics Modeling and Trajectory Prediction}

Physical dynamics models use material points, grids, particles, or graphs to represent phase-dependent states~\citep{jiang2016material,stomakhin2013material,macklin2016xpbd,modi2024simplicits,mantaflow,battaglia2016interaction,li2019dpi,sanchez2020learning,pfaff2021meshgraphnets,feng2025gaussiansplashing}. Fluid predictors incorporate continuous convolutions or equivariance~\citep{ummenhofer2020lagrangian,winchenbach2024sfbc,toshev2023learning}, while latent motion priors model non-rigid motion~\citep{tevet2023human,cao2024motion2vecsets,ji2025pomp}. Recent work learns physical parameters and dynamics from videos~\citep{garcia2025learningphysics,chen2025vid2sim,zhao2025masiv,fluidnexus2025,li2025trace,quan2026particlegs,perini2026windinet} or predicts scene and object trajectories~\citep{zhang2026moge4d,soraki2026objectforesight}; PhysCtrl and equation-based forecasting connect trajectory prediction to video generation~\citep{physctrl,feng2025physicsgroundedmotion}. HGATSolver, Fisale, and CFC explicitly couple distinct physical domains~\citep{zhang2026hgatsolver,tao2026fisale,dou2025cfc}. PAVG instead learns fixed-identity point trajectories for video control across single-phase motion and solid--gas interaction.

\section{Preliminaries}
\label{sec:preliminaries}

Solids and gases share momentum conservation but have distinct dynamics, motivating phase-specific models~\citep{baumgarten2021coupled,tao2026fisale}.
Their motions can be coupled through forces transmitted across the solid--gas interface.

\subsection{Solid Dynamics}
\label{sec:solid-dynamics}

Solid stress depends on deformation and history~\citep{masi2023evolution,raj2026history}.
With deformation gradient $\mathbf{F}_s$, internal variables $\boldsymbol{\zeta}_s$, and material properties $\boldsymbol{\phi}_s$, the constitutive laws are
\begin{equation}
\begin{aligned}
\frac{D\mathbf{F}_s}{Dt}
&=(\nabla\mathbf{u}_s)\mathbf{F}_s,\\
\boldsymbol{\sigma}_s
&=\mathcal{C}_s
(\mathbf{F}_s,\boldsymbol{\zeta}_s;\boldsymbol{\phi}_s),\\
\frac{D\boldsymbol{\zeta}_s}{Dt}
&=\mathcal{G}_s
(\mathbf{F}_s,\boldsymbol{\zeta}_s;\boldsymbol{\phi}_s).
\end{aligned}
\label{eq:solid-constitutive}
\end{equation}
Here, $\mathbf u_s$ is solid velocity, $D/Dt$ is the material derivative, $\mathcal C_s$ maps material state to stress $\boldsymbol{\sigma}_s$, and $\mathcal G_s$ governs internal history.
Elastic deformation is reversible; plastic changes persist.
Rigid motion satisfies $\mathbf{F}_s^{\mathsf T}\mathbf{F}_s=\mathbf I$, allowing translation and rotation.
Contact forces restrict admissible motion.

\subsection{Gas Dynamics}
\label{sec:gas-dynamics}

Gas motion is described by velocity field $\mathbf U_g$ and transported concentration $c_g$, rather than deformation-dependent stress ~\citep{li2026smokesvd}.
For nearly incompressible flow, ${\nabla\!\cdot\mathbf U_g\approx0}$, and
\begin{equation}
\begin{aligned}
\partial_t c_g
+\mathbf{U}_g\!\cdot\nabla c_g
&=\kappa_c\nabla^2c_g,\\
\varrho_0\left(
\partial_t\mathbf{U}_g
+(\mathbf{U}_g\!\cdot\nabla)\mathbf{U}_g
\right)
&=-\nabla p+\mu\nabla^2\mathbf{U}_g\\
&\quad
+\varrho_0\mathbf{b}_g(c_g)
+\mathbf{f}^{\mathrm{int}}_g,
\end{aligned}
\label{eq:gas-dynamics}
\end{equation}
where $\varrho_0$ is reference density, $\kappa_c$ is concentration diffusivity, and $\mu$ is dynamic viscosity.
Pressure $p$ enforces incompressibility.
Body acceleration $\mathbf b_g$ includes buoyancy; $\mathbf f_g^{\mathrm{int}}$ is interphase force density.
Gas evolution depends on transport, pressure, and dissipation, not material memory.

\subsection{Solid--Gas Interaction}
\label{sec:prelim-interaction}

We consider mechanical interaction at a solid--gas interface $\Gamma_{sg}$.
Neglecting interfacial inertia and other surface forces, momentum balance gives equal and opposite tractions~\citep{liu2018unified}:
\begin{equation}
\begin{aligned}
\boldsymbol{\sigma}_s\mathbf n_s
+\boldsymbol{\sigma}_g\mathbf n_g&=\mathbf0,
\quad \mathbf x\in\Gamma_{sg},\\
\boldsymbol{\sigma}_g&=-p\mathbf I+\boldsymbol{\tau}_g.
\end{aligned}
\label{eq:interface-traction-balance}
\end{equation}
Here, $\mathbf n_s$ and $\mathbf n_g$ are outward unit normals, and $\boldsymbol{\tau}_g$ is viscous stress.
\revnew{Pressure and viscous stresses couple solid motion and deformation with gas flow.}

Detailed dynamical models and their simulation implementations are provided in Supplementary Sec.~A.

\section{Methods}
\label{sec:methods}

\subsection{Overview}
\label{sec:method-overview}

Given selected regions of the input image $I_0$, we initialize phase-specific 3D point clouds $\mathbf P^0$.
PAVG ($\mathcal T_\theta$) predicts their 3D trajectories from physical controls $\mathbf a^{\mathrm{traj}}$ and phase/material conditions $\boldsymbol\phi$.
The projected tracks, $I_0$, and text prompt $y$ guide a pretrained video generator $\mathcal G_\psi$ (Supplementary Sec.~C.3):
\begin{equation}
\begin{aligned}
\widehat{\mathcal P}^{1:F_{\mathrm{phys}}}
&=\mathcal T_\theta\!\left(\mathbf P^0,\mathbf a^{\mathrm{traj}},\boldsymbol\phi\right),\\
\widehat{\mathcal V}_{1:F}
&=\mathcal G_\psi\!\left(I_0,y,\Pi\!\left(\widehat{\mathcal P}^{1:F_{\mathrm{phys}}}\right)\right).
\end{aligned}
\label{eq:video-problem}
\end{equation}
Here, $\Pi$ projects 3D trajectories to image-space tracks, and $F_{\mathrm{phys}}$ and $F$ are the physical and video frame counts; $\mathcal P$ denotes future 3D point positions.

\subsection{Phase-aware point-trajectory predictor}
\label{sec:phase-aware-trajectory}

\begin{figure*}[t]
\centering
\input{\pavgroot/figures/model_notation.tex}
\caption{\revnew{PAVG architecture: (a) dual-branch denoising; (b) gated CPI, with the gas update shown and phases swapped for the solid update; (c) spatial and temporal attention axes.}}
\label{fig:dual-stream}
\end{figure*}

\revnew{PAVG couples two phase-specific point-cloud diffusion branches~\citep{physctrl} through spatial and temporal cross-phase interaction (CPI; Figure~\ref{fig:dual-stream}).
Given phase regions and physical controls (e.g., material properties and applied forces), each branch retrieves context from the other through gated residual attention, preserving phase-specific dynamics while avoiding full joint attention over all points and frames.}

\paragraph{Phase-specific input encoding.}
For phase $q\in\{s,g\}$, the input comprises $N$ fixed-identity points, initial cloud $\mathbf P_q^0$, control $\mathbf a_q$, and material properties $\boldsymbol\phi_q$.
At diffusion step $\tau$, noisy trajectories $\mathcal P_{q,\tau}$ are encoded as
\begin{equation}
\begin{aligned}
(\mathbf H_q,\mathbf C_q)
&=\operatorname{Encode}_q(\mathcal P_{q,\tau},\mathbf c_q),\\
\mathbf c_q&=(\mathbf P_q^0,\mathbf a_q,\boldsymbol\phi_q).
\end{aligned}
\label{eq:branch-encoding}
\end{equation}
The encoder prepends the initial cloud and embeds positions as point features $\mathbf H_q$; force, material, and boundary descriptors form condition tokens $\mathbf C_q$.

\paragraph{Cross-phase interaction.}
Each of $L$ blocks applies spatial self-attention, spatial CPI, a feed-forward network, temporal self-attention, and temporal CPI.
Self-attention and feed-forward layers remain branch-specific, with PhysCtrl's residual connections and timestep-conditioned normalization; spatial attention and feed-forward layers also update condition tokens.
Both CPI modules use gated residual cross-attention (Figure~\ref{fig:dual-stream}b):
\begingroup
\ifshownewrev\color{newrevblue}\fi
\begin{equation}
\begin{aligned}
\mathbf H'_q&=\mathbf H_q+\tanh(\beta_{m,q})\\
&\quad\cdot\operatorname{Attn}_{m,q}\!\bigl(\underbrace{\vphantom{\mathbf H_q}\mathbf H_q}_{Q},\underbrace{\vphantom{\mathbf H_q}\mathbf H_r}_{K},\underbrace{\vphantom{\mathbf H_q}\mathbf H_r}_{V}\bigr),\\
(q,r)&\in\{(s,g),(g,s)\},\\
m&\in\{\mathrm{sp},\mathrm{temp}\}.
\end{aligned}
\label{eq:cross-attention}
\end{equation}
\endgroup
\revnew{where $\mathbf H_s$ and $\mathbf H_g$ denote solid and gas features, and $\beta_{m,q}$ is a learned scalar initialized to zero.}
\revold{Spatial CPI reads all solid points in the same frame; temporal CPI reads matching indices across frames after spatial mixing, not physical solid--gas correspondences.}
\revnew{Spatial CPI attends to all points of the other phase in the same frame. Temporal CPI attends across frames at matching point indices after spatial mixing; these indices do not imply physical solid--gas correspondences.}
\revold{CPI follows the directional local-transfer model (Supplementary Sec.~A.3, Eq.~S8), updates only gas features, and uses distinct parameters in each block to propagate features and condition tokens.}
\revnew{Each block has its own spatial and temporal CPI parameters. The updated phase features then continue through their respective branches.}

\paragraph{Trajectory denoising and output.}
\revnew{Branch-specific heads apply timestep-conditioned normalization and 3D projection, adding $\mathbf P_q^0$ to predicted offsets to obtain $\widehat{\mathcal P}_{q,\tau}$.
DDIM iteratively samples all physical frames jointly from Gaussian noise with fixed $\mathbf P_q^0$; single-phase inputs bypass CPI.}

\subsection{Training}
\label{sec:training-objectives}

\revnew{Stage~1 learns phase-specific motion priors; Stage~2 initializes PAVG from these priors and jointly learns single-phase motion and cross-phase interaction.}

\paragraph{Stage 1: Single-phase priors.}
For single-phase rollouts, we sample $\tau$ from a $K$-step schedule and $\boldsymbol\epsilon\sim\mathcal N(\mathbf0,\mathbf I)$, form $\mathcal P_{q,\tau}=\alpha_\tau\mathcal P_q+\sigma_\tau\boldsymbol\epsilon$, and predict clean positions $\widehat{\mathcal P}_q$ from $\mathbf c_q$ and $\tau$.
We supervise positions and frame-to-frame motion with
\begin{equation}
\begin{aligned}
\mathcal L_{\mathrm{diff}}^q
&=\mathbb E_{\mathcal P_q,\tau,\boldsymbol\epsilon}
\|\widehat{\mathcal P}_q-\mathcal P_q\|_2^2,\\
\mathcal L_{\mathrm{vel}}^q
&=\operatorname{mean}_f
\|\Delta\widehat{\mathbf P}_q^f-\Delta\mathbf P_q^f\|_2^2,
\end{aligned}
\label{eq:stage1-losses}
\end{equation}
where $\mathbf P_q^f$ is the frame-$f$ point cloud, $\Delta\mathbf P_q^f=\mathbf P_q^{f+1}-\mathbf P_q^f$, and $f=1,\ldots,F_{\mathrm{phys}}-1$.
For solids, we additionally use PhysCtrl's deformation and contact-plane penetration losses, $\mathcal L_{\mathrm{def}}$ and $\mathcal L_{\mathrm{floor}}$~\citep{physctrl}.

For gas, we impose MantaFlow's pressure-projection incompressibility constraint~\citep{mantaflow}.
Because targets are tracer trajectories rather than Eulerian grids, a locally fitted displacement Jacobian $\mathbf J_k$ yields
\begin{equation}
\nabla\!\cdot\mathbf U_g\approx0,
\qquad
d_k=\operatorname{tr}(\mathbf J_k)
\approx\delta t_f\,\nabla\!\cdot\mathbf U_g(\mathbf x_k),
\label{eq:training-incompressibility}
\end{equation}
where $\delta t_f$ is the stored-frame interval and $d_k,\widehat d_k$ come from simulated and predicted trajectories.
The divergence loss is
\begin{equation}
\mathcal{L}_{\mathrm{div}}
=\operatorname{mean}_{k\in\Omega_{\mathrm{div}}}
\operatorname{Huber}\!\left(
\frac{\widehat{d}_k-d_k}{\sigma_d}\right).
\label{eq:divergence-loss}
\end{equation}
Here, $\Omega_{\mathrm{div}}$ contains valid neighborhoods and $\sigma_d$ is the lower-bounded batchwise RMS of $d_k$.
Matching $d_k$, rather than zero, accounts for discretization and sampling.
With $\ell_q=\mathcal L_{\mathrm{diff}}^q+ \lambda_{\mathrm{vel}}\mathcal L_{\mathrm{vel}}^q$, the complete Stage~1 objectives are
\begin{equation}
\begin{aligned}
\mathcal{L}_{\mathrm{stage1}}^s
&=\ell_s+\lambda_{\mathrm{def}}\mathcal{L}_{\mathrm{def}}
+\lambda_{\mathrm{floor}}\mathcal{L}_{\mathrm{floor}},\\
\mathcal{L}_{\mathrm{stage1}}^g
&=\ell_g+\lambda_{\mathrm{div}}\mathcal{L}_{\mathrm{div}}.
\end{aligned}
\label{eq:stage1-total-loss}
\end{equation}
Each $\lambda$ weights its corresponding objective.

\paragraph{Stage 2: Interaction learning.}
Stage~2 mixes single-solid, single-gas, and paired solid--gas trajectories, applying Eq.~\ref{eq:stage1-total-loss} to present branches and independent noise to paired targets at a shared $\tau$.
\revnew{We jointly optimize both phase-specific branches and the CPI modules.}
\revold{For paired trajectories, the local coupling model (Supplementary Sec.~A.3) transfers solid-induced momentum to gas in the solid-affected region, and we supervise the corresponding velocity increment.}
\revnew{For paired trajectories, recorded solid-induced gas-velocity increments supervise local interactions (Supplementary Sec.~A.3).}

For gas point $p$ at frame $f$, predicted positions $\widehat{\mathbf x}_{g,p}^f$ give
\begin{equation}
\begin{aligned}
\widehat{\mathbf{v}}_{g,p}^{f}
&=\frac{\widehat{\mathbf{x}}_{g,p}^{f}
-\widehat{\mathbf{x}}_{g,p}^{f-1}}{\delta t_f},\\
\widehat{\Delta\mathbf{v}}_{g,p}^{f,\mathrm{total}}
&=\widehat{\mathbf{v}}_{g,p}^{f}
-\widehat{\mathbf{v}}_{g,p}^{f-1}.
\end{aligned}
\label{eq:derived-velocity-transfer}
\end{equation}
Using the recorded free-gas increment $\Delta\mathbf v_{g,p}^{f,\mathrm{free}}$ (Supplementary Sec.~C.2), we define
\begin{equation}
\widehat{\Delta\mathbf v}_{g,p}^{f,s\rightarrow g}
:=\widehat{\Delta\mathbf v}_{g,p}^{f,\mathrm{total}}
-\Delta\mathbf v_{g,p}^{f,\mathrm{free}}.
\label{eq:predicted-solid-induced-increment}
\end{equation}
With $\widehat{\mathbf x}_{g,p}^0=\mathbf x_{g,p}^0$ and $\widehat{\mathbf v}_{g,p}^0=\mathbf a_{g,p}$, we impose in the interaction region
\begin{equation}
\begin{aligned}
\mathcal{L}_{\mathrm{coupling}}
&=\operatorname{mean}_{(f,p)\in\Omega_{\mathrm{cpl}}}\\
&\quad
\operatorname{Huber}\!\left(
            \frac{\widehat{\Delta\mathbf{v}}_{g,p}^{f,s\rightarrow g}
            -\Delta\mathbf{v}_{g,p}^{f,s\rightarrow g}}
{\sigma_{\mathrm{cpl}}}\right),
\end{aligned}
\label{eq:coupling-loss}
\end{equation}
The recorded transfer defines valid pairs $\Omega_{\mathrm{cpl}}$; its lower-bounded batchwise componentwise RMS gives $\sigma_{\mathrm{cpl}}$.
The decomposition is given in Supplementary Sec.~C.2.
Stage~2 combines the complete phase-specific objectives:
\begin{equation}
\begin{aligned}
\mathcal{L}_{\mathrm{traj}}
&=\frac{\sum_{q\in\{s,g\}}w_q\mathcal L_{\mathrm{stage1}}^q}{w_s+w_g},\\
\mathcal{L}_{\mathrm{stage2}}
&=\mathcal{L}_{\mathrm{traj}}
 +\lambda_{\mathrm{coupling}}\mathcal{L}_{\mathrm{coupling}}.
\end{aligned}
\label{eq:stage2-loss}
\end{equation}
In each batch, $\mathcal L_{\mathrm{stage1}}^q$ averages valid phase-$q$ samples, with $w_q=0$ when phase $q$ is absent.

\section{Experiments}
\label{sec:experiments}

\begin{figure*}[t]
\centering
\includegraphics[width=\textwidth]{\pavgroot/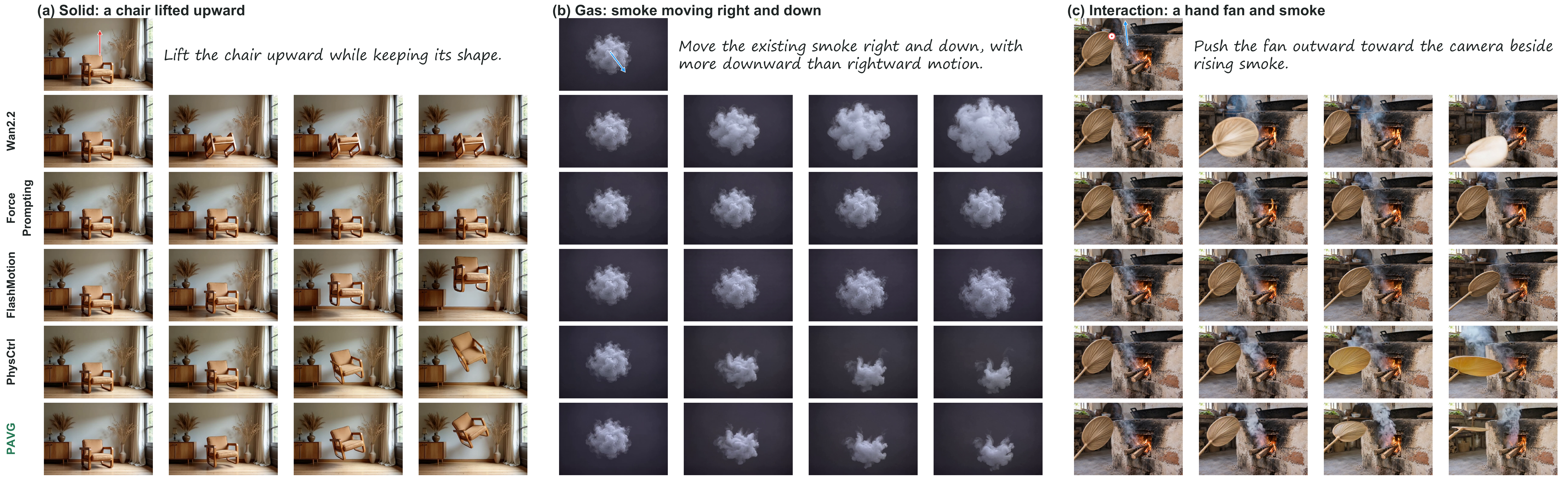}
\caption{\revnew{Qualitative video results of PAVG and baseline methods.}}
\label{fig:video-qualitative}
\end{figure*}


\subsection{Benchmarks}
\label{sec:exp-setup}
\paragraph{Video generation.}
\revnew{We evaluate 32 scenes: eight solid cases from PhysCtrl~\citep{physctrl}, eight gas cases, and sixteen solid--gas interactions.
Solid cases cover elastic, rigid, plasticine, and sand objects.
Gas cases cover smoke, fine dust, and water mist.
Interactions cover objects moving through dust, mist, steam, and clouds.
All 11-method video comparisons and PAVG/PhysCtrl point-cloud animations are available in the supplementary media (Section~1).
Construction and conditions are detailed in Supplementary Sec.~B.6.}

\paragraph{Trajectory prediction.}
\revnew{Our trajectory dataset and benchmark cover solid motion, gas motion, and their interactions; solid data follow PhysCtrl's generation procedure~\citep{physctrl} (Supplementary Sec.~B).}

\subsection{Motion Video Generation}
\label{sec:exp-video}

\paragraph{Baselines.}
We select baselines from \emph{general-purpose video generation} (HunyuanVideo-1.5, Wan2.2-I2V-A14B, and CogVideoX1.5-5B-I2V); \emph{motion-controlled generation} (DragAnything, ObjCtrl-2.5D, FlashMotion, and Wan-Move); and \emph{physics-controlled generation} (Force Prompting, StreamForce, and PhysCtrl).
Method-specific inputs are detailed in Supplementary Sec.~C.5.

\paragraph{\revnew{Quantitative and qualitative evaluation.}}
Four multimodal judges---GPT-4o, GPT-5.5, GPT-5.6-sol, and DeepSeek-v4.1-flash---evaluate all eleven methods on 32 scenes.
Each rates \emph{Semantic Adherence} (SA), \emph{Physical Commonsense} (PC), and \emph{Video Quality} (VQ) on a five-point Likert scale~\citep{bansal2024videophy,physctrl}.
\revnew{Evaluation protocols are detailed in Supplementary Sec.~C.5.}
Table~\ref{tab:video-quality} averages each metric equally over scenes and judges.
\revnew{PAVG ranks first in SA and VQ and ties with PhysCtrl for first in PC.}
\revnew{This limited margin likely stems from an evaluation mismatch. Reference-free MLLM judges lack physical ground truth and saturate on overall quality, while subtle gas effects are easily overshadowed by the foreground and background in generic metrics, as our qualitative examples illustrate in Figure~\ref{fig:video-qualitative}.}
\revnew{In Figure~\ref{fig:video-qualitative}, several baselines produce visually plausible gas and solid--gas videos, yet their motions deviate from the specified initial gas velocities and forces applied to solids. PAVG better follows these controls, preserving directed smoke transport and the gas response to the fan's motion.}
\revnew{Video scores for solid, gas, and solid--gas scenes are reported separately in Supplementary Table~S5.
Supplementary Fig.~S1 reports our user study.}

\revnew{To further verify this explanation and examine whether the generated motions obey physical laws, we provide point-cloud trajectory validations in the following section, which more directly reveal the advantages of our method.}


\begin{table}[t]
\centering
\caption{Motion video generation on the 32-scene benchmark.
Scores are averaged equally over scenes and four judges; higher is better.
\textbf{Bold}: best.}
\label{tab:video-quality}
\small
\setlength{\tabcolsep}{3pt}
\renewcommand{\arraystretch}{0.95}
\begin{tabular*}{\linewidth}{@{\extracolsep{\fill}}lccc@{}}
\toprule
Method & SA $\uparrow$ & PC $\uparrow$ & VQ $\uparrow$\\
\midrule
\multicolumn{4}{l}{\emph{General-purpose video generation}}\\
HunyuanVideo-1.5 & \revnum{3.04} & \revnum{3.06} & \revnum{3.66}\\
Wan2.2-I2V-A14B & \revnum{2.84} & \revnum{2.93} & \revnum{3.54}\\
CogVideoX1.5-5B-I2V & \revnum{2.17} & \revnum{2.34} & \revnum{2.95}\\
\midrule
\multicolumn{4}{l}{\emph{Motion-controlled generation}}\\
DragAnything & \revnum{2.27} & \revnum{2.41} & \revnum{2.72}\\
ObjCtrl-2.5D & \revnum{2.28} & \revnum{2.35} & \revnum{2.80}\\
FlashMotion & \revnum{3.06} & \revnum{3.16} & \revnum{3.56}\\
Wan-Move & \revnum{2.06} & \revnum{1.96} & \revnum{2.27}\\
\midrule
\multicolumn{4}{l}{\emph{Physics-controlled generation}}\\
Force Prompting & \revnum{2.02} & \revnum{2.21} & \revnum{2.81}\\
StreamForce & \revnum{1.88} & \revnum{2.05} & \revnum{2.72}\\
PhysCtrl & \revnum{3.52} & \revnum{\textbf{3.33}} & \revnum{3.70}\\
\midrule
\multicolumn{4}{l}{\emph{Ours}}\\
PAVG & \revnum{\textbf{3.55}} & \revnum{\textbf{3.33}} & \revnum{\textbf{3.71}}\\
\bottomrule
\end{tabular*}
\end{table}

\subsection{PAVG: Trajectory Prediction}
\label{sec:exp-trajectory}

\begin{figure*}[t]
\centering
\includegraphics[width=\textwidth]{\pavgroot/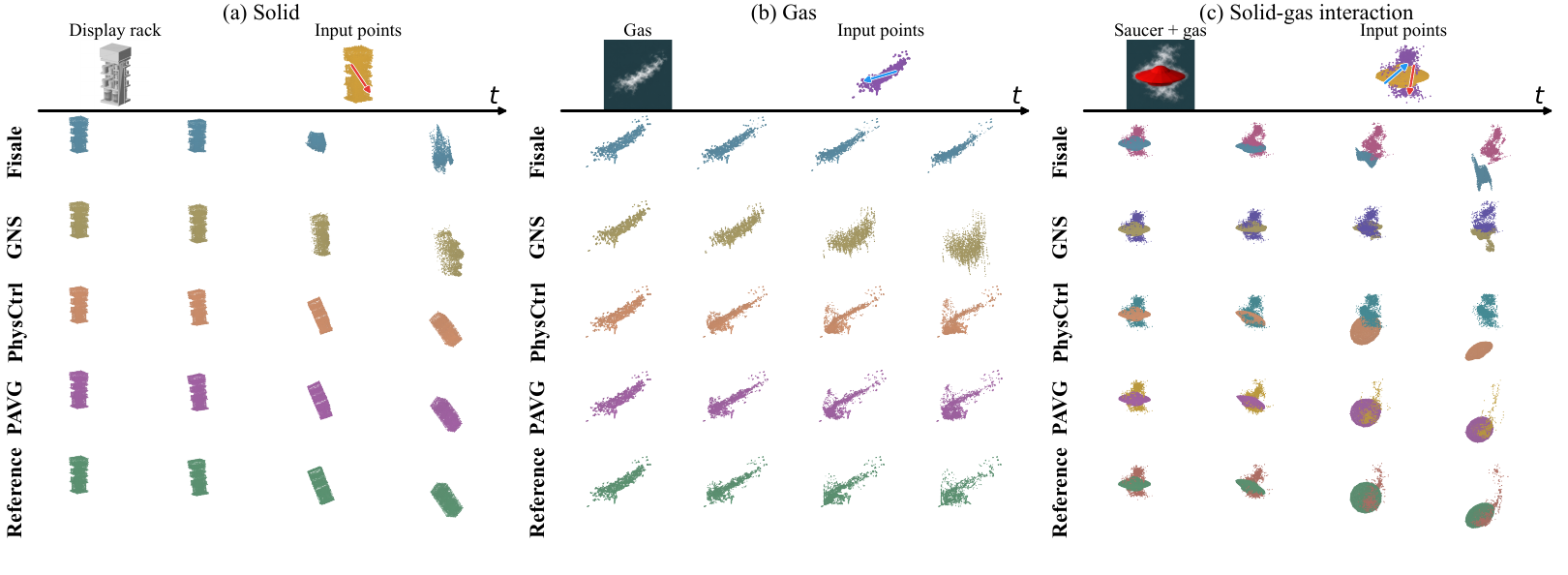}
\caption{\revnew{Trajectory comparison on a display rack, gas plume, and saucer--gas interaction.}}
\label{fig:trajectory-qualitative}
\end{figure*}

\begin{table*}[t]
\centering
\caption{Trajectory prediction across single-phase motion and solid--gas interaction.}
\label{tab:overall-trajectory}
\scriptsize
\setlength{\tabcolsep}{2.1pt}
\resizebox{\textwidth}{!}{%
\begin{tabular}{l@{\hskip 3pt}ccc@{\hskip 5pt}ccc@{\hskip 5pt}ccc@{\hskip 5pt}ccc}
\toprule
Method & \multicolumn{3}{c}{Overall$^*$} &
\multicolumn{3}{c}{Single-phase: Solid} &
\multicolumn{3}{c}{Single-phase: Gas} &
\multicolumn{3}{c}{Interaction: Solid--gas}\\
\cmidrule(lr){2-4}\cmidrule(lr){5-7}\cmidrule(lr){8-10}\cmidrule(lr){11-13}
& vIoU$\uparrow$ & CD$\downarrow$ & Corr$\downarrow$
& vIoU$\uparrow$ & CD$\downarrow$ & Corr$\downarrow$
& vIoU$\uparrow$ & CD$\downarrow$ & Corr$\downarrow$
& vIoU$\uparrow$ & CD$\downarrow$ & Corr$\downarrow$\\
\midrule
\multicolumn{13}{l}{\textbf{Gas-only}}\\
SFBC & \multicolumn{3}{c}{\textemdash} &
\multicolumn{3}{c}{\textemdash} &
0.483874 & 0.120906 & 0.274170 &
\multicolumn{3}{c}{\textemdash}\\
DLF & \multicolumn{3}{c}{\textemdash} &
\multicolumn{3}{c}{\textemdash} &
0.483505 & 0.053650 & 0.241004 &
\multicolumn{3}{c}{\textemdash}\\
SEGNN & \multicolumn{3}{c}{\textemdash} &
\multicolumn{3}{c}{\textemdash} &
0.561893 & 0.007060 & 0.176081 &
\multicolumn{3}{c}{\textemdash}\\
\midrule
\multicolumn{13}{l}{\textbf{Solid-only}}\\
Motion2VecSets & \multicolumn{3}{c}{\textemdash} &
0.141172 & 0.842913 & 0.656012 &
\multicolumn{3}{c}{\textemdash} &
\multicolumn{3}{c}{\textemdash}\\
MDM & \multicolumn{3}{c}{\textemdash} &
0.357344 & 0.097770 & 0.246086 &
\multicolumn{3}{c}{\textemdash} &
\multicolumn{3}{c}{\textemdash}\\
\midrule
\multicolumn{13}{l}{\textbf{Single-phase and interaction}}\\
GNS & 0.465241 & 0.027480 & 0.209879 &
0.422463 & 0.039714 & 0.163346 &
0.559694 & 0.007163 & 0.176219 &
0.439404 & 0.031521 & 0.249974\\
HGATSolver & 0.449227 & 0.068079 & 0.223303 &
0.306664 & 0.163993 & 0.309513 &
0.570428 & 0.006277 & 0.116932 &
0.459907 & 0.051023 & 0.233384\\
Fisale & 0.518143 & 0.036205 & 0.182453 &
0.478780 & 0.054490 & 0.144728 &
0.578805 & 0.005874 & 0.117675 &
0.507492 & 0.042229 & 0.233703\\
PhysCtrl & \revnum{0.625331} & \revnum{0.017875} & \revnum{0.124548} &
\revnum{0.613280} & \revnum{0.022733} & \revnum{0.071951} &
\revnum{0.783158} & \revnum{0.001423} & \revnum{0.065287} &
\revnum{0.552442} & \revnum{0.023672} & \revnum{0.180476}\\
PAVG & \textbf{0.705371} & \textbf{0.006663} & \textbf{0.059832} &
\textbf{0.627285} & \textbf{0.018528} & \textbf{0.066111} &
\textbf{0.793380} & \textbf{0.001353} & \textbf{0.061098} &
\textbf{0.700409} & \textbf{0.003384} & \textbf{0.056061}\\
\bottomrule
\end{tabular}
}%
\par\vspace{1mm}
\begin{minipage}{\linewidth}
\footnotesize
$^*$ Overall $O_m=\tfrac14 S_m+\tfrac14 G_m+\tfrac12 I_m$, where $S_m$, $G_m$, and $I_m$ are single-solid, single-gas, and interaction scores. Interaction metrics use concatenated solid and gas points. \textemdash: inapplicable; \textbf{bold}: best.
\end{minipage}
\end{table*}

\begin{figure*}[!t]
\centering
\includegraphics[width=\textwidth]{\pavgroot/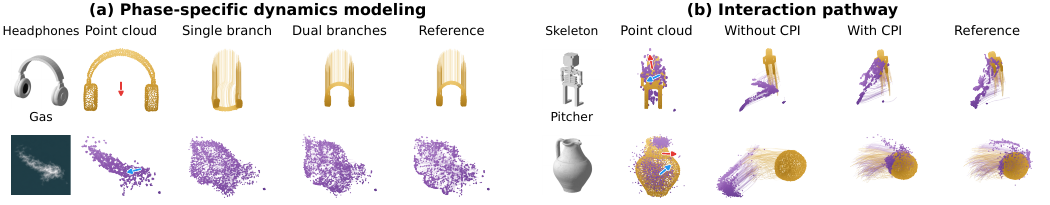}
\caption{\revnew{Qualitative ablations: (a) phase-specific modeling; (b) CPI with a skeleton figurine and a pitcher.}}
\label{fig:interaction-qualitative}
\end{figure*}

\begin{table*}[!t]
\begin{minipage}[t]{0.48\textwidth}
\centering
\caption{Phase-specific dynamics modeling.}
\label{tab:ablation-phase-dynamics}
\footnotesize
\setlength{\tabcolsep}{0.5pt}
\renewcommand{\arraystretch}{0.95}
\begin{tabular*}{\linewidth}{@{\extracolsep{\fill}}lcccccc@{}}
\toprule
& \multicolumn{3}{c}{Single branch} & \multicolumn{3}{c}{Dual branches (PAVG)}\\
\cmidrule(lr){2-4}\cmidrule(lr){5-7}
Setting & vIoU$\uparrow$ & CD$\downarrow$ & Corr$\downarrow$
& vIoU$\uparrow$ & CD$\downarrow$ & Corr$\downarrow$\\
\midrule
Overall & 0.625331 & 0.017875 & 0.124548
& \textbf{0.705371} & \textbf{0.006663} & \textbf{0.059832}\\
Single solid & 0.613280 & 0.022733 & 0.071951
& \textbf{0.627285} & \textbf{0.018528} & \textbf{0.066111}\\
Single gas & 0.783158 & 0.001423 & 0.065287
& \textbf{0.793380} & \textbf{0.001353} & \textbf{0.061098}\\
Interaction & 0.552442 & 0.023672 & 0.180476
& \textbf{0.700409} & \textbf{0.003384} & \textbf{0.056061}\\
\bottomrule
\end{tabular*}
\end{minipage}\hfill
\begin{minipage}[t]{0.48\textwidth}
\centering
\caption{CPI ablation on 100 interactions: joint (Overall) and phase-subset metrics.}
\label{tab:ablation-cross-attention}
\footnotesize
\setlength{\tabcolsep}{0.5pt}
\renewcommand{\arraystretch}{0.95}
\begin{tabular*}{\linewidth}{@{\extracolsep{\fill}}lcccccc@{}}
\toprule
& \multicolumn{3}{c}{PAVG (without CPI)} & \multicolumn{3}{c}{PAVG (with CPI)}\\
\cmidrule(lr){2-4}\cmidrule(lr){5-7}
Subset & vIoU$\uparrow$ & CD$\downarrow$ & Corr$\downarrow$
& vIoU$\uparrow$ & CD$\downarrow$ & Corr$\downarrow$\\
\midrule
Overall & 0.556162 & 0.024528 & 0.180640
& \textbf{0.700409} & \textbf{0.003384} & \textbf{0.056061}\\
Solid & 0.737668 & 0.010850 & 0.048697
& \textbf{0.751505} & \textbf{0.010130} & \textbf{0.045461}\\
Gas & 0.408280 & 0.115155 & 0.312583
& \textbf{0.651218} & \textbf{0.003325} & \textbf{0.066660}\\
\bottomrule
\end{tabular*}
\end{minipage}
\end{table*}

\begin{figure*}[!t]
\centering
\includegraphics[width=\textwidth]{\pavgroot/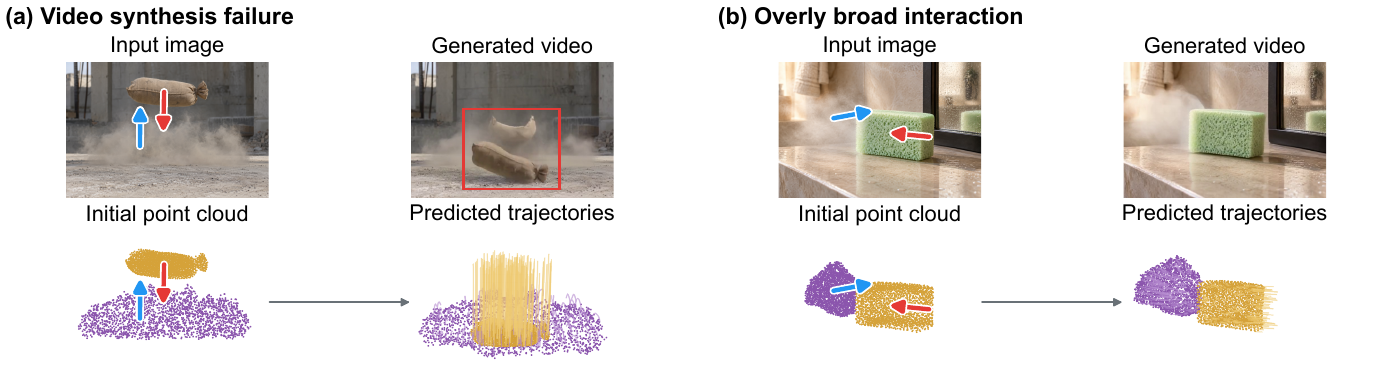}
\caption{\revnew{Failure cases: (a) two sandbags generated instead of one; (b) overly broad solid--gas influence.}}
\label{fig:pointcloud-failure-cases}
\end{figure*}

\paragraph{Baselines.}
\revold{We compare PhysCtrl, MDM, and Motion2VecSets for \emph{solid trajectory modeling}; SFBC, DLF, and SEGNN for \emph{fluid/gas trajectory modeling}; and GNS, HGATSolver, and Fisale for \emph{cross-phase interaction modeling}.}
\revnew{We train solid models (MDM, Motion2VecSets), gas models (SFBC, DLF, SEGNN), and models for both single-phase motion and interactions (PhysCtrl, GNS, HGATSolver, Fisale) on the corresponding trajectory subsets (Supplementary Secs.~C.1 and~C.4).}

\paragraph{Quantitative evaluation.}
We evaluate trajectories using voxel IoU (vIoU), Chamfer distance (CD), and correspondence-aware pointwise $L_2$ error (Corr)~\citep{physctrl,lei2022cadex,cao2024motion2vecsets}, averaged over future frames and test sequences.
Table~\ref{tab:overall-trajectory} shows that PAVG achieves the best overall and solid--gas interaction results on all three metrics.

SFBC and DLF predict gas dynamics autoregressively rather than modeling full fixed-identity trajectories\revnew{, whereas MDM and Motion2VecSets provide motion-prior baselines on solids}.
\revold{For interactions, GNS, HGATSolver, and Fisale use autoregressive updates with particle-type message passing, heterogeneous attention, and latent ALE grids, respectively.}
\revnew{PhysCtrl, GNS, HGATSolver, and Fisale cover single-phase and interaction modeling using a shared trajectory predictor, particle-type message passing, heterogeneous attention, and latent ALE grids, respectively~\citep{physctrl,sanchez2020learning,zhang2026hgatsolver,tao2026fisale}.}
PAVG instead jointly denoises all future frames with phase-specific spatiotemporal Transformers and \revold{solid-to-gas CPI}\revnew{CPI}, preserving within-phase dynamics and cross-phase influence in a shared trajectory space.

\paragraph{Qualitative evaluation.}
\revnew{Figure~\ref{fig:trajectory-qualitative} compares a display rack under force, a transported gas plume, and a saucer-shaped object moving through gas.
Fisale and GNS introduce excessive solid deformation and fail to reproduce the gas plume's curved transport.
For interactions, PhysCtrl separates gas from the moving solid, while PAVG better preserves relative motion and solid shape.}

\subsection{Ablation Studies}
\label{sec:ablations}

\paragraph{\revnew{Phase-specific dynamics modeling.}}
\revnew{We compare dual-branch PAVG with a single-branch predictor sharing parameters across phases (Table~\ref{tab:ablation-phase-dynamics}). Separate branches better preserve phase-specific dynamics.}

\paragraph{Interaction pathway ablation.}
\revnew{On the interaction test set, removing CPI markedly reduces gas trajectory accuracy while affecting solids less (Table~\ref{tab:ablation-cross-attention}), consistent with the stronger influence of solids on gas in these scenes.}

\paragraph{\revnew{Qualitative comparison.}}
\revnew{Figure~\ref{fig:interaction-qualitative}(a) compares falling headphones and gas transport; Figure~\ref{fig:interaction-qualitative}(b) examines a skeleton figurine and a pitcher moving through gas.
The single-branch predictor introduces solid motion and shape deviations.
CPI better preserves gas distributions and motion relative to solids; removing it produces displaced or distorted gas trajectories.
Ablation training and evaluation settings are detailed in Supplementary Sec.~C.4.}

\section{\revnew{Discussion}}
\label{sec:discussion}

\paragraph{\revnew{Failure cases.}}
\revnew{Although the predicted solid trajectory preserves a single coherent sandbag, the downstream video generator produces two sandbags (Fig.~\ref{fig:pointcloud-failure-cases}(a)).
Coordinate normalization can lead PAVG to overestimate the spatial range of solid--gas interactions in larger scenes (Fig.~\ref{fig:pointcloud-failure-cases}(b)).}

\paragraph{\revnew{Extension to liquids.}}
\revnew{Retaining the backbone, we train a liquid predictor with density/viscosity conditioning and contact-height supervision.
Its trajectories capture gravity-driven settling and material-dependent spreading in water, milk, edible oil, and honey, demonstrating effective extension to liquid motion prediction.
GT-paired results are provided in Supplementary Sec.~D.2 and supplementary media, Section~3.
A liquid branch can readily extend the design to solid--gas--liquid interaction prediction.}

\section{Conclusion}
In this work, we introduced PAVG (\emph{Phase-Aware Video Generation}), a framework for physics-grounded dynamics and interactions. PAVG explicitly separates the distinct dynamics of solids and gases through a dual-branch trajectory predictor, while spatiotemporal cross-attention captures the physical coupling between phases. By learning physical trajectory priors from the proposed large-scale trajectory corpus, PAVG enables efficient and controllable video generation under external physical interventions. Experiments across single-phase and solid-gas interaction scenarios show consistent improvements in motion adherence, physical plausibility, and visual quality. We hope this work provides a step toward more general physics-grounded video generation models capable of reasoning about heterogeneous physical dynamics and their interactions.

\revnew{PAVG currently relies on user-specified physical conditions.
Future work could use vision-language models to estimate initial physical conditions from images and text.}

\clearpage
\typeout{PAVG-MAIN-TEXT-PAGES: \number\numexpr\value{page}-1\relax}

\section*{LLM Usage Statement}

In this work, we used \revnew{\texttt{gpt-image-2.5}} to generate the initial scene images used as inputs for the comparative video-generation experiments and used generative AI tools to evaluate generated videos with the fixed four-judge protocol described in Section~\ref{sec:exp-video} and Supplementary Sec.~C.5. We also used \revnew{\texttt{gpt-6.1sol}} to create or edit software code, edit the manuscript to improve its readability and wording, and correct LaTeX formatting. We did not use generative AI tools for literature retrieval or discovery, generating synthetic datasets, developing theoretical models or conceptual frameworks, formulating or proving mathematical claims, or designing or providing feedback on the research methodology or experiments. The initial scene images were experiment inputs and were not used as a synthetic dataset; the model ratings were collected under a predefined evaluation protocol and did not determine the method, experimental protocol, or conclusions. We reviewed all AI-assisted work: the authors manually checked the generated images, reviewed and tested all code modifications, checked the evaluation protocol and resulting scores, and checked the manuscript, figures, tables, equations, and reported results. The authors made the final decisions about the method, experimental protocol, and conclusions, and take full responsibility for the final content of this work, including all text, code, images, figures, results, and other artifacts produced with AI assistance.

We used GPT-4o, GPT-5.5, GPT-5.6-sol, and DeepSeek-v4.1-flash as multimodal judges to score generated videos for Semantic Adherence (SA), Physical Commonsense (PC), and Video Quality (VQ).

{\small
\IfFileExists{CVPR/references.bib}{%
  \bibliographystyle{CVPR/ieeenat_fullname}
  \bibliography{CVPR/references}
}{%
  \bibliographystyle{ieeenat_fullname}
  \bibliography{references}
}
}

\clearpage
\typeout{PAVG-MAIN-PDF-PAGES: \number\numexpr\value{page}-1\relax}
\onecolumn
\appendix
\setcounter{page}{1}
\setcounter{equation}{0}
\setcounter{figure}{0}
\setcounter{table}{0}
\renewcommand{\theequation}{S\arabic{equation}}
\renewcommand{\thefigure}{S\arabic{figure}}
\renewcommand{\thetable}{S\arabic{table}}
\renewcommand{\theHequation}{supp.\arabic{equation}}
\renewcommand{\theHfigure}{supp.\arabic{figure}}
\renewcommand{\theHtable}{supp.\arabic{table}}
\begin{center}
{\Large\bfseries \thetitle\par}
\vspace{0.5em}
{\Large Supplementary Material\par}
\end{center}
\raggedbottom

\subsection*{Reproducibility statement}

We provide a detailed description of PAVG and its phase-aware point-trajectory predictor in main-paper Section~4, including the model interface, cross-phase interaction modules, and two-stage training objectives. Main-paper Section~5 specifies the benchmark composition, baselines, evaluation metrics, and ablation protocols. Solid, gas, and solid--gas simulation dynamics are detailed in Sec.~\ref{app:dynamics-simulation}; trajectory generation, data partitions, and video benchmark construction are described in Sec.~\ref{app:benchmark-construction}; and architecture, optimization, initialization, trajectory-evaluation, and video-scoring details are provided in Sec.~\ref{app:training-implementation}, including the fixed data splits, reported random seeds, sampling settings, and evaluation prompts. An anonymized implementation, configuration files, and supplementary experiment artifacts will be provided with the submission to facilitate independent verification, with a public release planned upon publication.

\section{Dynamics and Simulation}
\label{app:dynamics-simulation}

This section expands the phase-specific dynamics in main-paper Section~3 and describes the solvers used to generate trajectory supervision.
Solid simulation follows the MPM and rigid-body formulation used by PhysCtrl~\citep{physctrl,jiang2016material}, whereas gas simulation uses MantaFlow~\citep{mantaflow} with an additional local coupling update for interaction scenes.
The simulators are used for data generation, not for online video synthesis.

\subsection{Solid Dynamics and Simulation}
\label{app:solid-dynamics-simulation}

\paragraph{Continuum description.}
For a material coordinate $\mathbf X$, the deformation map $\boldsymbol\chi(\mathbf X,t)$ gives position $\mathbf x$ and deformation gradient $\mathbf F=\partial\boldsymbol\chi/\partial\mathbf X$.
Mass and momentum balance take the form
\begin{equation}
\frac{D\varrho_s}{Dt}+\varrho_s\nabla\!\cdot\mathbf u_s=0,
\qquad
\varrho_s\frac{D\mathbf u_s}{Dt}
=\nabla\!\cdot\boldsymbol\sigma_s+\mathbf f_s^{\mathrm{ext}},
\label{eq:appendix-solid-balance}
\end{equation}
where $\mathbf f_s^{\mathrm{ext}}$ is external force per unit volume, including gravity when enabled.
For an elastic strain-energy density $\Psi$, the first Piola stress $\mathbf P$ and Cauchy stress satisfy
\begin{equation}
\mathbf P=\frac{\partial\Psi}{\partial\mathbf F},
\qquad
\boldsymbol\sigma_s=J^{-1}\mathbf P\mathbf F^{\mathsf T},
\qquad J=\det\mathbf F.
\label{eq:appendix-solid-stress}
\end{equation}
Material parameters determine $\Psi$; plastic updates additionally retain irreversible changes through the material state in main-paper Eq.~1.
Thus, the same balance equations support distinct elastic, plasticine, and sand responses through their constitutive and return-mapping rules~\citep{jiang2016material}.

\paragraph{Material point discretization.}
MPM stores positions, velocities, volumes, and deformation states on material points and evaluates their interactions through a background grid~\citep{physctrl,jiang2016material}.
Let $w_{ip}=N_i(\mathbf x_p^n)$ be the interpolation weight from point $p$ to grid node $i$, $\mathbf C_p$ its affine velocity matrix, and $\Delta t$ a simulation substep.
Particle-to-grid transfer and momentum update:
\begin{equation}
\begin{aligned}
m_i&=\sum_p w_{ip}m_p,\\
m_i\mathbf u_i^n
&=\sum_p w_{ip}m_p
\left[\mathbf v_p^n+\mathbf C_p^n(\mathbf x_i-\mathbf x_p^n)\right],\\
\mathbf u_i^*
&=\mathbf u_i^n+\frac{\Delta t}{m_i}
\left[\mathbf f_i^{\mathrm{ext}}
-\sum_p V_p^0\mathbf P_p^n(\mathbf F_p^n)^{\mathsf T}\nabla N_i(\mathbf x_p^n)\right].
\end{aligned}
\label{eq:appendix-mpm-p2g}
\end{equation}
Here, $V_p^0$ is the reference point volume; updates use only grid nodes with nonzero mass.
Boundary and contact conditions modify $\mathbf u_i^*$ to obtain $\mathbf u_i^{n+1}$.
Grid-to-particle transfer then gives
\begin{equation}
\begin{aligned}
\mathbf v_p^{n+1}&=\sum_i w_{ip}\mathbf u_i^{n+1},\\
\mathbf x_p^{n+1}&=\mathbf x_p^n+\Delta t\mathbf v_p^{n+1},\\
\mathbf F_p^{\mathrm{trial}}
&=\left[\mathbf I+\Delta t\sum_i
\mathbf u_i^{n+1}\nabla N_i(\mathbf x_p^n)^{\mathsf T}\right]\mathbf F_p^n,\\
\mathbf F_p^{n+1}&=\mathcal R_{\mathrm{mat}}(\mathbf F_p^{\mathrm{trial}}).
\end{aligned}
\label{eq:appendix-mpm-g2p}
\end{equation}
The material-specific return map $\mathcal R_{\mathrm{mat}}$ is the identity for the elastic update and enforces the selected yield response for plastic materials.
The affine matrix is updated from the same grid velocities, preserving the local velocity variation needed by the next transfer.
These substeps are distinct from the more sparsely stored trajectory frames used for learning.

\paragraph{Rigid motion and contact.}
Rigid examples use a rigid-body solver rather than a deformable material with an artificially large stiffness.
For center of mass $\mathbf x_c$, rotation $\mathbf R$, angular velocity $\boldsymbol\omega$, mass $m$, and world-space inertia $\mathbf I_w$, the motion obeys
\begin{equation}
\begin{aligned}
m\dot{\mathbf v}_c&=\mathbf f^{\mathrm{ext}}+\mathbf f^{\mathrm{contact}},
&\dot{\mathbf x}_c&=\mathbf v_c,\\
\frac{d}{dt}(\mathbf I_w\boldsymbol\omega)
&=\boldsymbol\tau^{\mathrm{ext}}+\boldsymbol\tau^{\mathrm{contact}},
&\dot{\mathbf R}&=[\boldsymbol\omega]_\times\mathbf R.
\end{aligned}
\label{eq:appendix-rigid-dynamics}
\end{equation}
Here, $[\boldsymbol\omega]_\times\mathbf a=\boldsymbol\omega\times\mathbf a$.
Fixed body-space points follow $\mathbf x_p(t)=\mathbf x_c(t)+\mathbf R(t)\mathbf r_p$, so their pairwise distances remain unchanged.
For both rigid and deformable examples, floor contact supplies reaction forces or velocity corrections that prevent penetration.
The material, force, and sampling choices used to construct the solid data are given in Sec.~\ref{app:solid-data}.

\subsection{Gas Dynamics and Simulation}
\label{app:gas-dynamics-simulation}

\paragraph{Incompressible flow approximation.}
MantaFlow represents gas velocity on a staggered MAC grid and transports a scalar concentration field~\citep{mantaflow}.
In our setup, $c_g$ describes the visible smoke or aerosol concentration, rather than a spatially varying density in a compressible equation of state.
The reference carrier density $\varrho_0$ is held constant, and pressure enforces the approximate divergence-free condition in main-paper Eq.~2.
This model describes transport and buoyancy without resolving acoustic waves or phase changes.

\paragraph{Advection, forcing, and pressure projection.}
Let $\mathcal A_{\Delta t}(\mathbf U,q)$ denote semi-Lagrangian transport of a field $q$ by velocity $\mathbf U$ over one step.
The implementation uses second-order advection for concentration and velocity, followed by buoyancy and pressure projection~\citep{mantaflow}.
Its principal operators can be written as
\begin{equation}
\begin{aligned}
c_g^{n+1}
&=\max(1-\gamma\Delta t,0)\,
\mathcal A_{\Delta t}(\mathbf U_g^n,c_g^n),\\
\mathbf U_g^*
&=\mathcal A_{\Delta t}(\mathbf U_g^n,\mathbf U_g^n)
+\Delta t\,\mathbf b_g(c_g^{n+1}),\\
\nabla^2p^{n+1}
&=\frac{\varrho_0}{\Delta t}\nabla\!\cdot\mathbf U_g^*,
\qquad
\mathbf U_g^{n+1}
=\mathbf U_g^*-\frac{\Delta t}{\varrho_0}\nabla p^{n+1}.
\end{aligned}
\label{eq:appendix-gas-projection}
\end{equation}
The last two expressions give the continuum form of the discrete projection; the solver incorporates the domain boundary conditions.
The coefficient $\gamma$ controls concentration decay and is not a molecular diffusion coefficient.
Our scene implementation does not add separate explicit viscosity or concentration-diffusion solves; smoothing can also arise from advection and interpolation.
Thus, the general viscous and diffusive terms in main-paper Section~3.2 should not be interpreted as independently tuned, nonzero terms in every generated sequence.
Wall conditions are applied around the pressure solve, while open boundaries are specified by the scene configuration.
In interaction scenes, the transfer described below is inserted before an additional projection of the recorded velocity field.

\paragraph{Tracer trajectories.}
Learning targets are persistent tracers from the initial gas support, not Eulerian grid cells.
For interpolated velocity $\mathcal I[\mathbf U_g]$, continuous and discrete tracer motion obey
\begin{equation}
\dot{\mathbf x}_{g,p}=\mathcal I[\mathbf U_g](\mathbf x_{g,p}),
\qquad
\mathbf x_{g,p}^{n+1}
=\mathbf x_{g,p}^{n}+\Delta t\,
\mathcal I[\mathbf U_g^n](\mathbf x_{g,p}^{n}).
\label{eq:appendix-gas-tracers}
\end{equation}
The exporter preserves tracer identities, records concentration weights and alive flags, and applies the configured obstacle and domain-boundary handling.
Tracers leaving the valid domain are marked inactive rather than replaced by new identities.

\subsection{Solid--Gas Interaction Dynamics and Simulation}
\label{app:simulation-coupling}

\paragraph{Interface conditions.}
Main-paper Equation~3 is the dynamic interface condition.
For an impermeable interface with no slip, the kinematic condition is $\mathbf u_s=\mathbf U_g$ on $\Gamma_{sg}$ ~\citep{liu2018unified}.
For an incompressible Newtonian gas, $\boldsymbol{\tau}_g=\mu(\nabla\mathbf U_g+\nabla\mathbf U_g^{\mathsf T})$.
The traction $\boldsymbol{\sigma}_g\mathbf n_g$ is the force per unit area exerted on the gas by the solid; integrating it over the interface gives the resultant force.
Its counterpart on the solid has the opposite sign.
In a boundary-based formulation, this exchange enters through interface tractions; in a distributed-force formulation, it is represented by interphase force densities, such as $\mathbf f_g^{\mathrm{int}}$ in main-paper Eq.~2.
These describe one exchange and must not be counted twice.

\paragraph{Local transfer approximation.}
Our data-generation implementation uses a localized acceleration model rather than resolving the full two-sided interface problem.
Let $\mathbf r_{sg}=\mathcal E_s[\mathbf u_s]-\mathbf U_g$, where $\mathcal E_s$ extends solid velocity into the nearby gas region.
For outward solid normal $\mathbf n_{sg}$, the tangential component is $\mathbf r^t_{sg}=(\mathbf I-\mathbf n_{sg}\mathbf n_{sg}^{\mathsf T}) \mathbf r_{sg}$.
The local transfer rule is
\begin{equation}
\mathbf{a}_{s\rightarrow g}
=w_{sg}
\left[
k_n[\mathbf{r}_{sg}\!\cdot\mathbf{n}_{sg}]_+\mathbf{n}_{sg}
+k_t\mathbf{r}^{t}_{sg}
\right],
\label{eq:interaction-response}
\end{equation}
where $[u]_+=\max(u,0)$, $k_n,k_t$ set transfer strengths, and the nonnegative weight $w_{sg}(\mathbf x,t)$ localizes interaction near the solid.
Under the one-way approximation, $\mathbf f_g^{\mathrm{int}}=\varrho_0\mathbf a_{s\rightarrow g}$ and $\mathbf f_s^{\mathrm{int}}\approx\mathbf0$.
This is an implementation-specific approximation, not an exact enforcement of main-paper Eq.~3.
MantaFlow handles gas evolution and pressure projection; the coupling update uses replayed solid points.
The effective solid velocity equals its simulated velocity except near floor contact, where a positive frame-to-frame jump in the floor-normal component is added to account for the collision impulse.

The local weight in Eq.~\ref{eq:interaction-response} is
\begin{equation}
w_{sg}(\mathbf{x},t)
=\mathbf{1}\!\left[c_g(\mathbf{x},t)>c_{\min}\right]
\left[1-\frac{d(\mathbf{x},\Omega_s(t))}{R}\right]_+^2,
\label{eq:interaction-weight}
\end{equation}
where $\Omega_s(t)$ is the solid region, $d$ is the distance to it, $R$ is the influence radius, and $c_{\min}$ is the gas-concentration threshold.
Thus, weights decay quadratically to zero at $R$.
After accumulating and normalizing point contributions, we cap the increment norm.
In cells occupied by solid points, an occupancy-weighted correction blends this increment toward the difference between the local solid and gas velocities.
The resulting update is applied before pressure projection and tracer transport.
The recorded $\Delta\mathbf{v}_{g,p}^{f,s\rightarrow g}$ samples this applied update at the gas tracers; it is not the difference between separately simulated coupled and uncoupled trajectories.

\paragraph{Discrete coupling update.}
The implementation evaluates local stencils around the replayed solid points, using the direction from each solid point to a gas-grid location as a local normal surrogate.
For grid location $i$ and solid point $j$, let $w_{ij}$ be the density-gated distance weight and $\mathbf a_{ij}$ the bracketed relative-velocity response in Eq.~\ref{eq:interaction-response}.
Overlapping point contributions produce
\begin{equation}
\begin{aligned}
\Delta\mathbf U_i^{\mathrm{pen}}
&=\operatorname{clip}_{v_{\max}}
\left(\frac{\Delta t\sum_j w_{ij}\mathbf a_{ij}}
{\max(1,\sum_j w_{ij})}\right),\\
\Delta\mathbf U_i
&=(1-o_i)\Delta\mathbf U_i^{\mathrm{pen}}
+o_i(\overline{\mathbf u}_{s,i}-\mathbf U_{g,i}),
\end{aligned}
\label{eq:appendix-discrete-coupling}
\end{equation}
where $\operatorname{clip}_{v_{\max}}$ caps the vector norm, $o_i\in[0,1]$ is particle occupancy, and $\overline{\mathbf u}_{s,i}$ is the occupancy-weighted solid velocity.
The second expression drives occupied cells toward the local solid velocity without deleting their concentration.
The stateful coupling configuration uses $k_n=28$, $k_t=5$, particle radius $0.012$, influence radius $R=0.045$, $c_{\min}=10^{-4}$, and $v_{\max}=0.45$ in the normalized simulation coordinates.
For recorded floor-collision events, an additional localized upward and radial velocity increment transfers the pre-impact normal impulse to nearby gas.
The resulting field is projected before export, while the applied pre-projection transfer is retained as a separate diagnostic for the interaction constraint.

\section{Data Generation and Benchmarks}
\label{app:benchmark-construction}

\subsection{Trajectory Representation}
\label{app:trajectory-representation}

Each trajectory sample contains an initial point cloud, numerical motion conditions, and future positions with fixed point identities.
We retain $N=2{,}048$ points for each present phase and supervise 24 future physical frames.
Prior work finds that this point budget provides comparable or better video-generation performance than other evaluated settings, including larger point sets~\citep{physctrl}.
For paired samples, the solid and gas have separate point identities and share the same coordinate transform and sampled times.
Point indices are not correspondences between phases.
Phase-presence flags distinguish single-phase samples from interactions, and gas alive flags identify invalid tracer entries.
The stored initial state is used as a condition rather than a prediction target.
The training loaders select future stored-frame indices $1,3,\ldots,47$; physical velocities use the associated elapsed times, including the shorter first interval where applicable.
Simulation coordinates, model normalization, and image projection are separate transforms and are not interpreted as calibrated camera-space SI measurements.

\subsection{Solid Trajectory Generation}
\label{app:solid-data}

We generate solid trajectories following PhysCtrl's asset curation and simulation methodology~\citep{physctrl}, rather than importing its complete trajectory dataset.
The source geometry consists of ObjaverseXL assets selected by TRELLIS-500K.
Objects are normalized, randomly oriented, and sampled with farthest-point sampling to obtain the control-point representation.
Material assignments produce elastic, plasticine, sand, and rigid examples using the solvers in Sec.~\ref{app:solid-dynamics-simulation}.
For deformable materials, the generation recipe samples $\log_{10}E$ uniformly from $[4,7]$ and Poisson's ratio $\nu$ from $[0.05,0.45]$.
Elastic examples include localized-force and gravity-driven motion, whereas plasticine, sand, and rigid examples form gravity-driven subsets.
Localized-force conditions record the application point, affected-point mask, and force vector; gravity examples record the gravity flag and floor height.
Rigid trajectories apply the simulated object transform to fixed body-space sample points.
Exports contain positions, velocities, material labels, physical conditions, and, where defined by MPM, deformation gradients, affine velocity matrices, and point volumes.

\subsection{Gas Trajectory Generation}
\label{app:gas-data}

Gas initialization draws on OpenVDB/JangaFX and WildSmoke assets to obtain a spatial concentration distribution and an initial tracer cloud.
These assets supply initial states; subsequent motion is generated under the prescribed controls using the MantaFlow procedure in Sec.~\ref{app:gas-dynamics-simulation}.
We vary the initial gas support, orientation, and velocity condition while retaining 2,048 tracer identities.
The random-motion generation recipe samples an isotropic direction and a speed from a Gaussian with mean $2.0$ and standard deviation $0.4$, truncated to $[0.8,3.2]$ in its normalized coordinates.
The global initial-motion descriptor is supplemented by spatial expansion, swirl, and smooth perturbations to form a nonuniform initial velocity field.
It is not a continuously applied external force.
The corresponding recipe uses a $28\times36\times28$ gas grid and 48 evolution intervals of $0.05$, with the initial state stored separately.
Continuous source emission and imposed domain wind are disabled in this single-gas recipe, so the rollout evolves the existing gas support.
The exporter records positions, sampled velocities, concentration weights, and validity flags before the learning-time frame selection.

\subsection{Interaction Trajectory Generation}
\label{app:interaction-data}

Interaction examples combine a simulated solid rollout with a gas initial state in a shared spatial domain.
The solid positions and velocities are replayed at the gas solver's recorded times, while the initial gas support is aligned relative to the solid or its contact plane.
MantaFlow advances the gas with the local transfer and collision updates in Sec.~\ref{app:simulation-coupling}.
Both point streams are exported together with their physical conditions and phase-presence flags.
Additional gas diagnostics record the total velocity change, the applied solid-induced velocity increment, and tracer validity.
These diagnostics provide the supervision region and scale used by the interaction constraint; they are not obtained by subtracting two independently simulated trajectories.
The solid and gas streams of one interaction remain together when assigning that sample to a dataset partition.

\subsection{Dataset Partitions and Benchmark}
\label{app:trajectory-benchmark}

The corpus comprises five solid subsets, one gas subset, and one interaction subset.
We use fixed training, validation, and test manifests, with 100 validation and 100 test trajectories reserved from each subset.
The benchmark consists of the seven held-out test subsets: 500 solid, 100 gas, and 100 solid--gas interaction sequences.
Validation samples are used for development, and the test membership is kept fixed across comparisons and ablations.
Single-phase baselines use the corresponding solid or gas partitions, while interaction-capable predictors use the combined partitions.
All methods share the same test point identities, physical-frame selection, and coordinate normalization.
Subset-level training and test counts are reported in Table~\ref{tab:datasets}.

\begin{table}[htbp]
\centering
\caption{Training datasets and held-out trajectory test sets.
The force condition denotes drag for solid sequences and initial velocity for gas sequences.}
\label{tab:datasets}
\begin{tabular}{llrr}
\toprule
Dataset & Simulator & Train & Test\\
\midrule
\multicolumn{4}{l}{\textbf{Solid}}\\
\quad elastic-force & MPM & 150,201 & 100\\
\quad elastic-gravity & MPM & 94,038 & 100\\
\quad plasticine-gravity & MPM & 71,356 & 100\\
\quad sand-gravity & MPM & 95,414 & 100\\
\quad rigid-gravity & rigid body & 97,091 & 100\\
\addlinespace
\multicolumn{4}{l}{\textbf{Gas}}\\
\quad gas-force & MantaFlow & 99,800 & 100\\
\addlinespace
\multicolumn{4}{l}{\textbf{Interaction}}\\
\quad solid-gas-force & MPM+MantaFlow & 99,792 & 100\\
\bottomrule
\end{tabular}
\end{table}

\subsection{Video Generation Benchmark}
\label{app:video-benchmark}

The 32-scene benchmark comprises eight solid, eight gas, and sixteen solid--gas interaction cases.
The solid cases use initial images and physical conditions from the released PhysCtrl examples~\citep{physctrl}, covering elastic, rigid, plasticine, and sand motion.
These examples provide initial-state inputs rather than ground-truth future trajectories.
\revnew{We construct the gas and interaction cases, with coverage summarized in Table~\ref{tab:video-scene-coverage}.
The gas-only set contains two smoke, three fine-dust, and three water-mist cases.
Interaction cases include a tennis ball moving horizontally into court dust, other balls and a sandbag moving through dust, steam or haze around a lifted pot lid, sponges and cushions disturbing mist or dust, and aircraft moving through clouds.}

\begin{table}[htbp]
\centering
\caption{\revnew{Coverage of the 32 video scenes.}}
\label{tab:video-scene-coverage}
\small
\begin{tabular}{@{}llr@{}}
\toprule
Subset & Appearance or interaction & Cases\\
\midrule
Gas & Smoke & 2\\
 & Airborne fine dust & 3\\
 & Water mist & 3\\
\addlinespace
Solid & Elastic, rigid, plasticine, and sand objects & 8\\
\addlinespace
Interaction & Balls moving through dust & 4\\
 & Lifted lids with steam or haze & 3\\
 & Sponges/cushions with mist or dust & 5\\
 & Aircraft moving through clouds & 3\\
 & Falling sandbag with rising dust & 1\\
\midrule
Total & & 32\\
\bottomrule
\end{tabular}
\end{table}

Each scene specifies an initial image, selected phase regions, numerical motion conditions, and a textual motion description.
Solid conditions specify material properties and applied forces or gravity; gas conditions specify initial motion rather than a sustained force.
These conditions are prescribed rather than estimated from the input image.
Scene identifiers link the clean input image, phase masks, numerical conditions, and generation prompt, and these inputs are fixed before comparison.
Methods receive the conditions through their supported interfaces, as described in main-paper Section~5.2.
For PAVG, the selected regions initialize phase-specific point clouds, whose predicted trajectories guide video synthesis through the implementation in Sec.~\ref{app:initialization-rendering}.
\revnew{Videos and point-cloud files are available in the supplementary media (Section~1).}

\section{Training and Implementation Details}
\label{app:training-implementation}

\subsection{Architecture and Optimization}
\label{app:architecture-optimization}

Each PAVG branch uses eight spatiotemporal Transformer blocks, a 256-dimensional latent space, and four attention heads.
The solid and gas branches have independent parameters and are connected by the spatial and temporal CPI modules defined in main-paper Section~4.2.
\revold{The interaction gates are initialized to zero, and solid context is detached when read by the gas branch.}
\revnew{The interaction gates are initialized to zero, with independent spatial and temporal cross-attention in each direction and gradients flowing through both branches.}
Inputs include the initial geometry, pointwise initial velocity and action mask, point volumes, and force, material, and boundary descriptors.
Material descriptors distinguish the four solid categories and gas; padded fields for absent phases do not create an additional motion target.

Stage~1 trains the two single-phase priors separately, and Stage~2 initializes the respective branches from these priors and trains on single-phase and paired samples.
For the main model, each Stage~1 predictor and the Stage~2 model use 60K optimizer updates with an effective batch size of 32.
The shared configuration and stage-specific batch layout are summarized in Table~\ref{tab:appendix-training}.
The diffusion objective predicts clean positions using a 1,000-step DDPM schedule with sample clipping disabled.
Trajectory inference uses the corresponding DDIM schedule with 25 denoising steps.
Checkpoints and data manifests are fixed for each reported comparison; test data are not used as additional conditioning during sampling.

\begin{table}[htbp]
\centering
\caption{PAVG architecture and optimization settings.}
\label{tab:appendix-training}
\small
\begin{tabular}{ll}
\toprule
Setting & Value\\
\midrule
Blocks / latent dimension / attention heads & 8 / 256 / 4 per branch\\
Points / future frames & 2,048 per phase / 24\\
Optimizer & AdamW, $\beta=(0.9,0.999)$, $\epsilon=10^{-8}$\\
Learning rate / weight decay & $10^{-4}$ / $10^{-2}$\\
Schedule / warmup & Cosine decay / 100 updates\\
Gradient clipping / precision & Norm 1.0 / bfloat16\\
Updates / effective batch & 60K per stage / 32\\
Stage 1 solid: GPUs $\times$ local batch $\times$ accumulation & $8\times1\times4$\\
Stage 1 gas: GPUs $\times$ local batch $\times$ accumulation & $8\times2\times2$\\
Stage 2: GPUs $\times$ local batch $\times$ accumulation & $16\times1\times2$\\
Diffusion training / inference steps & 1,000 / 25\\
\bottomrule
\end{tabular}
\end{table}

In the phase-specific training configurations, the velocity, deformation, floor, and gas-divergence weights are $1$, $10^{-3}$, $1$, and $0.05$, respectively, with each physical term applied to its applicable data.
The main Stage~2 model retains these phase-specific terms and weights, uses equal solid and gas branch weights, and adds a coupling coefficient of $0.05$.
Baseline-specific architectures and budgets are described in Sec.~\ref{app:trajectory-comparison}.

\subsection{Physical Constraints}
\label{app:physical-constraints}

\paragraph{Solid deformation and floor constraints.}
Following PhysCtrl~\citep{physctrl}, the deformation constraint reconstructs grid velocities from predicted point motion and recorded material state, then compares the updated and simulated deformation gradients:
\begin{equation}
\widetilde{\mathbf F}_p^{f+1}
=\left[\mathbf I+\delta t_f\sum_i
\widehat{\mathbf u}_i^{f+1}\nabla N_i(\widehat{\mathbf x}_p^f)^{\mathsf T}\right]
\mathbf F_p^f,
\label{eq:appendix-deformation-constraint}
\end{equation}
where the reference $\mathbf F_p^f$ and affine state are provided by the MPM data, and $\widehat{\mathbf u}_i^{f+1}$ is obtained by transferring the predicted point velocities to the grid.
Unlike the simulation step in Eq.~\ref{eq:appendix-mpm-p2g}, this supervision estimates the kinematic deformation update from trajectory motion rather than running a full stress-driven rollout during training.
It is applied only to samples with the required solid-state fields.
For a horizontal contact plane of height $h$ and upward $y$ coordinate, the penetration penalty is
\begin{equation}
\mathcal L_{\mathrm{floor}}
=\operatorname{mean}_{f,p}
\left[\max(h-\widehat x_{s,p,y}^f,0)\right]^2.
\label{eq:appendix-floor-constraint}
\end{equation}

\paragraph{Trajectory-level gas supervision.}
The training files contain point positions and interaction diagnostics, but not MantaFlow's Eulerian density, pressure, or velocity grids.
Consequently, all differentiable predicted quantities below are constructed from the predicted point positions.

\paragraph{Point-cloud divergence proxy.}
Let $\mathcal N(k)$ be center $k$'s nearest-neighbor set in the initial cloud.
For stored interval $f$, define the frame-to-frame and relative displacements for neighbor $j$:
\begin{equation}
\begin{aligned}
\mathbf u_p^f&=\mathbf x_p^{f+1}-\mathbf x_p^f,\\
\delta\mathbf x_{kj}^f&=\mathbf x_j^f-\mathbf x_k^f,\\
\delta\mathbf u_{kj}^f&=\mathbf u_j^f-\mathbf u_k^f.
\end{aligned}
\label{eq:appendix-relative-displacement}
\end{equation}
The same definitions are applied to the predicted positions, using hats.
We fit a local linear displacement map with ridge regularization,
\begin{align}
\mathbf J_k^f
&=\arg\min_{\mathbf J}
\sum_{j\in\mathcal N(k)}
\left\|\delta\mathbf u_{kj}^f-\mathbf J\delta\mathbf x_{kj}^f\right\|_2^2
 +\lambda_{\mathrm{reg}}\|\mathbf J\|_F^2, \nonumber\\
\widehat{\mathbf J}_k^f
&=\arg\min_{\mathbf J}
\sum_{j\in\mathcal N(k)}
\left\|\widehat{\delta\mathbf u}_{kj}^f-\mathbf J\widehat{\delta\mathbf x}_{kj}^f\right\|_2^2
 +\lambda_{\mathrm{reg}}\|\mathbf J\|_F^2.
\label{eq:appendix-local-jacobian}
\end{align}
The simulated and predicted divergence proxies are the traces
\begin{equation}
d_k^f=\operatorname{tr}(\mathbf J_k^f),
\qquad
\widehat d_k^f=\operatorname{tr}(\widehat{\mathbf J}_k^f).
\label{eq:appendix-divergence-trace}
\end{equation}
In the main text, the batch and frame indices are suppressed and $k$ denotes one center--time neighborhood.
In the implementation, at most 256 centers are selected uniformly from the input cloud, each uses 8 nearest neighbors, and $\lambda_{\mathrm{reg}}=10^{-3}$.
A center--time pair belongs to $\Omega_{\mathrm{div}}$ only when the center and at least three of its neighbors are valid in both the predicted and simulated trajectories.
With the simulated values used as the reference, the normalization is
\begin{equation}
\begin{aligned}
\sigma_d
&=\max\!\left(
\sqrt{\frac{1}{|\Omega_{\mathrm{div}}|}
\sum_{(b,f,k)\in\Omega_{\mathrm{div}}}
\left(d_{b,k}^f\right)^2},
\sigma_{d,\min}\right),\\
\sigma_{d,\min}&=1.0.
\end{aligned}
\label{eq:appendix-divergence-scale}
\end{equation}
The scale is recomputed for each batch and detached from the computation graph.
Since the fitted quantity uses displacement rather than velocity, it is proportional to $\Delta t\,\nabla\!\cdot\mathbf v$ for a fixed physical frame interval; its zero condition is therefore the same incompressibility proxy.

\paragraph{Solid-induced gas transfer scale.}
For a sampled interaction interval, the duration is $\delta t_f=\Delta t_{\mathrm{int}}m_f$, where $m_f$ is the number of raw simulation steps aggregated into that interval.
The predicted gas velocity is computed only from the model trajectory, initialized by the observed state:
\begin{equation}
\begin{aligned}
\widehat{\mathbf v}_{g,p}^f
&=\frac{\widehat{\mathbf x}_{g,p}^f-\widehat{\mathbf x}_{g,p}^{f-1}}
{\delta t_f},\\
\widehat{\mathbf x}_{g,p}^{0}&=\mathbf x_{g,p}^{0},
\quad
\widehat{\mathbf v}_{g,p}^{0}=\mathbf v_{g,p}^{0,\mathrm{init}}.
\end{aligned}
\label{eq:appendix-predicted-velocity}
\end{equation}
The subsequent predicted total velocity change is obtained by subtracting the previous predicted velocity, using the initial velocity only for the first interval.
The existing interaction files provide the total gas velocity change and the solid-induced transfer increment.
The loader converts both raw fields to the same sampled-interval representation, then defines
\begin{equation}
\begin{aligned}
\Delta\mathbf v_{g,p}^{f,\mathrm{free}}
&=\Delta\mathbf v_{g,p}^{f,\mathrm{total}}
-\Delta\mathbf v_{g,p}^{f,s\rightarrow g},\\
\widehat{\Delta\mathbf v}_{g,p}^{f,s\rightarrow g}
&=\widehat{\Delta\mathbf v}_{g,p}^{f,\mathrm{total}}
-\Delta\mathbf v_{g,p}^{f,\mathrm{free}}.
\end{aligned}
\label{eq:appendix-transfer-decomposition}
\end{equation}
The valid set $\Omega_{\mathrm{cpl}}$ contains gas point--interval pairs whose gas tracer is alive, whose interaction sample is valid, for which both phase streams are present, and whose simulated transfer magnitude is at least $10^{-5}$.
We normalize by the batchwise componentwise RMS of simulated transfer,
\begin{equation}
\begin{aligned}
\sigma_{\mathrm{cpl}}
&=\max\Biggl(
\sqrt{\frac{1}{3|\Omega_{\mathrm{cpl}}|}
\sum_{(b,f,p)\in\Omega_{\mathrm{cpl}}}
\left\|\Delta\mathbf v_{g,p}^{f,s\rightarrow g}\right\|_2^2},\\
&\hphantom{=\max\Biggl(}\sigma_{\mathrm{cpl},\min}\Biggr),\\
\sigma_{\mathrm{cpl},\min}&=10^{-4}.
\end{aligned}
\label{eq:appendix-coupling-scale}
\end{equation}
This scale is also detached from the computation graph.
Substituting Eq.~\ref{eq:appendix-transfer-decomposition} gives:
\begin{equation}
\widehat{\Delta\mathbf v}_{g,p}^{f,s\rightarrow g}
-\Delta\mathbf v_{g,p}^{f,s\rightarrow g}
=\widehat{\Delta\mathbf v}_{g,p}^{f,\mathrm{total}}
-\Delta\mathbf v_{g,p}^{f,\mathrm{total}}.
\label{eq:appendix-transfer-equivalence}
\end{equation}
Thus, main-paper Eq.~13 is algebraically identical to the implemented objective.
The recorded transfer selects and normalizes the supervision region; the optimized residual matches total velocity increments, not an independently inferred causal transfer.

\subsection{Image-to-Video Implementation}
\label{app:initialization-rendering}

\paragraph{Step 1: Phase-specific initialization.}
The video benchmark uses reviewed solid and gas masks to construct 2.5D point clouds in a shared image-aligned coordinate system.
For sampled pixel $(u_p,v_p)$, image-space center $(c_u,c_v)$, and scale $s$, the proxy coordinates are
\begin{equation}
\mathbf x_p^0
=\left[(u_p-c_u)/s,\ (c_v-v_p)/s,\ z_p\right]^{\mathsf T},
\label{eq:appendix-mask-initialization}
\end{equation}
where $z_p$ is a small depth perturbation used to give the proxy finite thickness.
The same center and scale are used for both phases in an interaction scene, preserving their image-space separation.
We sample 2,048 points per present phase and retain the inverse transform for projection.
Solid conditions specify material properties, initial velocity, force support, and the contact plane when applicable.
Gas conditions contain a prescribed global initial-motion descriptor and pointwise initial velocities.
Where a nonuniform initial field is required, the initializer maps a training-only gas initial state to the target support and prescribed direction; it does not use future trajectories from that sample.
These numerical conditions are not estimated from the single image.

A reconstruction-based initialization is also supported for selected solid regions: \texttt{rembg} and SAM~\citep{kirillov2023segment} isolate the object, SV3D~\citep{voleti2024sv3d} provides additional views, and LGM~\citep{tang2024lgm} reconstructs a representation from which points are sampled.
This optional path is distinct from the mask-derived benchmark initialization, whose results do not establish full volumetric gas reconstruction.

\paragraph{Step 2: Point-trajectory prediction.}
We encode the initial point clouds and numerical conditions using the same phase ordering and normalization as the training inputs.
PAVG samples all 24 future frames jointly, starting from Gaussian noise and applying 25 DDIM updates.
The video-benchmark trajectory runs use a fixed sampling seed of 42 and guidance scale of 1.0.
Single-phase inputs activate the corresponding branch; interaction inputs activate both branches and the learned interaction pathway.
The initial cloud remains fixed throughout denoising, and the returned coordinates are clean absolute positions rather than Eulerian fields.
Neither the text prompt nor future ground-truth motion is supplied to PAVG.

\paragraph{Step 3: Projection and video generation.}
For the mask-derived initialization, the stored inverse transform maps predicted positions back to pixel coordinates, with model depth retained in the track representation.
The reconstruction-based path instead uses its camera projection.
The current DaS adapter constructs 49 track frames from the initial cloud and the 24 forecasts: it inserts midpoints between successive forecasts and repeats the last endpoint.
This is a temporal control encoding, not a new physical simulation or an assertion that every video backbone follows the same physical clock.
Image-space control anchors inside each phase mask are assigned to nearby projected points; anchors outside the moving regions remain fixed, and out-of-frame tracks are marked invisible.
The two phase streams are merged into one tracking condition rather than synthesized as separate videos.
Pretrained DaS~\citep{gu2025das}, using the \texttt{cogshader5B} checkpoint, receives the clean initial image, merged tracking condition, and scene text.
The video-synthesis components are frozen and are not optimized during PAVG training.
The 32-scene video evaluation samples each frozen generated clip as described in Sec.~\ref{app:video-evaluation}; export FPS does not redefine the trajectory solver's stored-frame intervals.

\subsection{Trajectory Evaluation}
\label{app:trajectory-comparison}

\paragraph{Baseline adaptation.}
\revold{The gas-only comparison assigns SFBC, DLF, and SEGNN to gas training and test data; the solid-only comparison assigns PhysCtrl, MDM, and Motion2VecSets to solid training and test data.}
\revnew{The gas-only comparison assigns SFBC, DLF, and SEGNN to gas training and test data; the solid-only comparison assigns MDM and Motion2VecSets to solid training and test data. The task-adapted PhysCtrl predictor is trained on the exposure-matched schedule in Sec.~\ref{app:ablation-settings} and is evaluated on all benchmark subsets.}
\revold{The single-phase rows refer to phase-specific training, rather than subset scores from a combined model.}
\revnew{Gas-only and solid-only baseline rows denote phase-specific training; single-phase scores for unified and interaction-capable models are evaluated on their corresponding test subsets.}
For the solid-only comparison, MDM and Motion2VecSets are adapted to predict point trajectories from initial geometry and physical conditions, following the baseline setup in PhysCtrl~\citep{physctrl}.
The overall comparison evaluates the single-solid, single-gas, and paired solid--gas sets, with GNS, HGATSolver, and Fisale trained on their union under the mixed protocol.
The comparison fixes the available initial geometry and physical conditions and excludes ground-truth future states from inference inputs.
The resulting comparison concerns task-adapted implementations under fixed data splits and metrics, not the original methods in their native problem settings.

Baselines retain method-specific architectures and training budgets.
\revnew{The task-adapted PhysCtrl baseline uses eight Transformer blocks, 256 latent dimensions, and four attention heads, with 2,048 points per present phase. At test time, the same checkpoint predicts each phase separately from its own initial state and physical conditions, using 2,048 points per forward pass.}
The MDM configuration uses eight layers, 256 latent dimensions, and 60K updates, while Motion2VecSets uses 60K autoencoder updates followed by 60K diffusion updates, with a global batch size of 32 in both stages.
HGATSolver is trained for 60K updates with a global batch size of 16 and seed 42, and its checkpoint is selected by validation performance.
Fisale is trained for 60K updates with a global batch size of 24 and seed 42, and its checkpoint is selected by validation performance.
Autoregressive predictors roll out from the initial state without teacher-forcing future targets, while sequence predictors output the evaluation horizon jointly.

\paragraph{Metrics and aggregation.}
For predicted and reference clouds $\widehat P,P$ at a future frame, define
\begin{equation}
\begin{aligned}
\operatorname{CD}(\widehat P,P)
&=\frac{1}{|\widehat P|}\sum_{\widehat{\mathbf x}\in\widehat P}
\min_{\mathbf x\in P}\|\widehat{\mathbf x}-\mathbf x\|_2^2\\
&\quad+\frac{1}{|P|}\sum_{\mathbf x\in P}
\min_{\widehat{\mathbf x}\in\widehat P}\|\mathbf x-\widehat{\mathbf x}\|_2^2,\\
\operatorname{Corr}(\widehat P,P)
&=\frac{1}{N_P}\sum_{p=1}^{N_P}
\|\widehat{\mathbf x}_p-\mathbf x_p\|_2,
\end{aligned}
\label{eq:appendix-trajectory-metrics}
\end{equation}
where $N_P$ is the number of ordered points in the evaluated cloud.
CD sums the two directed squared-distance means without a factor of $1/2$, whereas Corr uses unsquared Euclidean distance and requires the fixed point identities.
For vIoU, points are voxelized on a grid with spacing $0.1$ over $[-1.5,1.5)^3$ in the evaluation coordinates, and occupied-cell intersection is divided by occupied-cell union.
Points outside this grid do not contribute occupied cells, and an empty union receives zero vIoU.
Each metric is averaged over the 24 future frames and then over samples in its evaluation subset.
The interaction score in main-paper Table~2 uses the concatenated solid and gas clouds, whereas the pathway ablation additionally evaluates the two streams separately.
Overall weights solid, gas, and interaction scores by $1/4$, $1/4$, and $1/2$, rather than by their frequencies among the 700 samples.

\paragraph{Ablation settings.}
\label{app:ablation-settings}
\revold{All ablation variants use fixed data splits, Base-capacity branches, and the same evaluation protocol.}
\revnew{All ablation variants use fixed data splits and the same metrics and sampling settings. The CPI comparison retains the same phase-specific branch capacity. PhysCtrl uses the task-adapted architecture specified in Sec.~\ref{app:trajectory-comparison}.}
Unless otherwise specified, each predictor is trained for 60K updates per stage with a global batch size of 32.
\revnew{Both shared predictors train from random initialization on the same schedule of 1.92 million scene exposures from each of the Stage~1 solid, Stage~1 gas, and Stage~2 sources. Each interaction is split into independent solid and gas examples of 2,048 points each, yielding 6,030,634 phase examples in total. Section~\ref{app:capacity-inference} describes the loss weights. At inference, each shared model predicts the two phases independently from their respective initial states and physical conditions. The two point sets are combined for interaction evaluation.}
\revnew{PhysCtrl uses global batch 32 for the first 2,500 updates and global batch 64 thereafter, completing the schedule in 95,479 updates. The nine-layer unified phase predictor uses global batch 64 throughout and completes the same schedule in 94,229 updates. We select each checkpoint by weighted Overall vIoU on the complete 700-sample validation set. This selects update 81,250 for PhysCtrl and 90,000 for the nine-layer predictor.}

\revold{The interaction-pathway comparison uses the same two-stage-trained model and disables only the solid-to-gas cross-attention updates at inference.}
\revnew{The interaction-pathway comparison separately trains PAVG with and without CPI from the same Stage~1 priors, keeping Stage~2 data, sample exposure, and losses fixed.}
\revnew{PAVG with CPI uses 60K updates at global batch 32; the variant without CPI uses 15K updates at global batch 128 (eight GPUs, local batch 16, accumulation 1). Both consume 1.92 million Stage~2 examples. Learning rate is $10^{-4}$ for both, and warmup, validation, and checkpoint intervals are scaled by sample exposure.}
\revold{Its quantitative evaluation covers the interaction subset of the trajectory benchmark, while its qualitative comparison shows three examples from that subset.}
\revnew{Quantitative evaluation covers joint, solid, and gas trajectories on the interaction subset; qualitative evaluation pairs both predictions with the same ground-truth cases.}
Initial conditions, sampling settings, and random seeds are matched between the two variants.

\subsection{Video Evaluation and Scoring Prompts}
\label{app:video-evaluation}

\paragraph{Inputs and evaluation procedure.}
All video methods receive the same clean initial image; the shared motion description is supplied only to methods with a text interface.
PAVG and the PhysCtrl baseline additionally receive numerical physical conditions for trajectory prediction, whereas Force Prompting and StreamForce receive controls through their native force interfaces.
The general-purpose baselines receive image and text conditions only.
For motion-controlled baselines, frozen trajectories from the PAVG control pipeline are converted into phase-wise masks and box trajectories for FlashMotion, point tracks and visibility for Wan-Move, phase-wise representative trajectories and masks for DragAnything, and a merged-entity representative trajectory and mask for ObjCtrl-2.5D.
For ObjCtrl-2.5D, the initial image is also used to estimate depth, and the merged trajectory is converted into camera poses using the released implementation.
We evaluate its merged-entity adapter rather than the multi-object extension demonstrated in its paper; this aggregation cannot preserve independent solid and gas motion.
These baselines evaluate alternative video synthesis backends under trajectory guidance.
Their frozen videos are retained when the PAVG and PhysCtrl videos are updated, so their control trajectories are not assumed to be identical to those in the updated PAVG release.
\revnew{The updated PAVG videos use revised solid-material assignments; retained methods keep the material settings used for their saved clips.}
Method-native negative prompts and inference settings are not treated as additional shared physical inputs.
The annotated condition image is supplied to the evaluator, not substituted for the clean generation input.

PAVG uses its full Stage~2 model.
\revold{For PhysCtrl, we use a checkpoint trained on the solid training split of our dataset for 60K updates with a task-adapted eight-layer, 256-dimensional Transformer and a global batch size of 32.}
\revnew{For PhysCtrl, we use the task-adapted predictor described in Sec.~\ref{app:trajectory-comparison}, trained on the same data and cumulative sample exposure as PAVG.}
Training uses AdamW with a learning rate of $10^{-4}$, weight decay of $10^{-2}$, cosine decay, and 100 warmup updates.
\revold{This PhysCtrl video checkpoint differs from the one used for trajectory evaluation in main-paper Table~2.}
\revnew{The same evaluated PhysCtrl checkpoint supplies the PhysCtrl trajectories for both trajectory and video comparisons.}
\revold{For solid--gas scenes, the PhysCtrl adapter jointly predicts 2,048 points (1,024 solid and 1,024 gas) in one pass, supplying gas velocity as per-point initial velocity rather than an external force.}
\revnew{For solid--gas scenes, the shared PhysCtrl model is called separately on 2,048 solid and 2,048 gas points. Each call excludes the other phase, and the outputs are combined only after sampling. Gas velocity is supplied as per-point initial velocity.}
\revold{This is an adaptation of the solid-trained predictor to gas and interaction inputs, not a gas-trained PhysCtrl model.}
\revold{Both methods use seed 42 for trajectory sampling and video generation and produce 49-frame clips at $720\times480$ resolution and 8 FPS.}
\revnew{Both methods use 25-step trajectory sampling with seed 42 and render 49 frames at $720\times480$ resolution with 50 DaS steps, guidance scale 6, video seed 20260910, and 24 FPS.}

\paragraph{Four-judge protocol and aggregation.}
Following PhysCtrl's reference-free criteria~\citep{physctrl}, GPT-4o, GPT-5.5, GPT-5.6-sol, and DeepSeek-v4.1-flash independently rate SA, PC, and VQ as integers from 1 to 5.
All four judges use the same frozen rubric on the same 32 scenes: eight solid, eight gas, and sixteen solid--gas cases.
For a clip with $F_v$ frames, the ten evaluation indices are $\operatorname{round}(j(F_v-1)/9)$ for $j=0,\ldots,9$, including the first and last frames.
Each video is represented by four vertical contact sheets containing samples 1--3, 4--6, 7--8, and 9--10, read from top to bottom.
Each request therefore contains 45 images: one annotated initial image and four sheets for each of eleven anonymous videos.
Red denotes solid force or gravity; cyan denotes initial gas velocity, not sustained wind.
The PNG sheets are preserved, while request images are encoded as JPEGs with a maximum side length of 768 pixels and quality 64, using the high-detail image setting.
Method order is shuffled separately for each scene with seed 20260924 and held fixed across judges; method identities and previous scores are not supplied.
The exact per-scene permutations, prompts, input hashes, and raw responses are retained with the evaluation records.

All 128 judge--scene requests are complete, yielding $4\times32\times11\times3=4{,}224$ scalar ratings.
For method $m$ and criterion $k$, we first compute the scene mean for each judge $j$ and then their equal-weight mean:
\begin{equation}
\bar{s}_{m,k}=\frac{1}{4}\sum_{j=1}^{4}
\left(\frac{1}{32}\sum_{c=1}^{32}s_{j,c,m,k}\right).
\end{equation}
SA, PC, and VQ are never combined into one score.
We use no intermediate rounding; displayed values are rounded half up to two decimals, and rankings use the unrounded means.
Incomplete or unparsable responses are not filled with zeros, and ratings from other video releases or rubric versions are not pooled.
Best values are marked in bold, with all ties retained.

\paragraph{Individual judge results.}
Each judge's mean scores over the same 32 scenes are reported in Table~\ref{tab:video-judges}.
The method labels and evaluated videos match main-paper Table~1.
Rankings vary by judge, so the aggregate lead does not imply that PAVG wins every metric for every judge.

\begin{table}[htbp]
\centering
\caption{Individual judge scores over the same 32 scenes. Higher is better for all metrics.
\textbf{Bold}: best within each judge and metric.}
\label{tab:video-judges}
\small
\setlength{\tabcolsep}{2pt}
\renewcommand{\arraystretch}{1.12}
\begin{tabular*}{\linewidth}{@{\extracolsep{\fill}}l*{12}{c}@{}}
\toprule
 & \multicolumn{3}{c}{GPT-4o}
 & \multicolumn{3}{c}{GPT-5.5}
 & \multicolumn{3}{c}{GPT-5.6-sol}
 & \multicolumn{3}{c}{\shortstack{DeepSeek-\\v4.1-flash}}\\
\cmidrule(lr){2-4}\cmidrule(lr){5-7}\cmidrule(lr){8-10}\cmidrule(lr){11-13}
Method & SA$\uparrow$ & PC$\uparrow$ & VQ$\uparrow$
 & SA$\uparrow$ & PC$\uparrow$ & VQ$\uparrow$
 & SA$\uparrow$ & PC$\uparrow$ & VQ$\uparrow$
 & SA$\uparrow$ & PC$\uparrow$ & VQ$\uparrow$\\
\midrule
HunyuanVideo-1.5 & \revnum{3.47} & \revnum{3.50} & \revnum{3.78} & \revnum{2.84} & \revnum{2.97} & \revnum{3.78} & \revnum{2.84} & \revnum{2.78} & \revnum{\textbf{3.53}} & \revnum{3.00} & \revnum{3.00} & \revnum{\textbf{3.56}}\\
Wan2.2-I2V-A14B & \revnum{3.19} & \revnum{3.22} & \revnum{3.59} & \revnum{2.81} & \revnum{2.88} & \revnum{3.69} & \revnum{2.47} & \revnum{2.72} & \revnum{3.41} & \revnum{2.88} & \revnum{2.91} & \revnum{3.47}\\
CogVideoX1.5-5B-I2V & \revnum{2.69} & \revnum{2.78} & \revnum{3.31} & \revnum{2.03} & \revnum{2.13} & \revnum{2.94} & \revnum{1.75} & \revnum{2.09} & \revnum{2.63} & \revnum{2.22} & \revnum{2.38} & \revnum{2.91}\\
DragAnything & \revnum{2.78} & \revnum{2.81} & \revnum{3.22} & \revnum{2.22} & \revnum{2.38} & \revnum{2.66} & \revnum{1.91} & \revnum{2.03} & \revnum{2.38} & \revnum{2.19} & \revnum{2.41} & \revnum{2.63}\\
ObjCtrl-2.5D & \revnum{2.59} & \revnum{2.59} & \revnum{3.09} & \revnum{2.31} & \revnum{2.25} & \revnum{2.81} & \revnum{2.00} & \revnum{2.09} & \revnum{2.63} & \revnum{2.22} & \revnum{2.47} & \revnum{2.69}\\
FlashMotion & \revnum{3.41} & \revnum{3.44} & \revnum{3.72} & \revnum{3.13} & \revnum{3.16} & \revnum{3.75} & \revnum{2.78} & \revnum{2.91} & \revnum{3.44} & \revnum{2.94} & \revnum{3.13} & \revnum{3.34}\\
Wan-Move & \revnum{2.72} & \revnum{2.69} & \revnum{3.16} & \revnum{1.97} & \revnum{1.81} & \revnum{2.13} & \revnum{1.84} & \revnum{1.72} & \revnum{1.81} & \revnum{1.72} & \revnum{1.63} & \revnum{1.97}\\
Force Prompting & \revnum{2.44} & \revnum{2.44} & \revnum{3.00} & \revnum{1.84} & \revnum{2.00} & \revnum{2.75} & \revnum{1.72} & \revnum{1.94} & \revnum{2.63} & \revnum{2.09} & \revnum{2.47} & \revnum{2.88}\\
StreamForce & \revnum{2.28} & \revnum{2.28} & \revnum{2.91} & \revnum{1.66} & \revnum{1.84} & \revnum{2.72} & \revnum{1.53} & \revnum{1.69} & \revnum{2.41} & \revnum{2.06} & \revnum{2.38} & \revnum{2.84}\\
PhysCtrl & \revnum{\textbf{3.75}} & \revnum{\textbf{3.78}} & \revnum{\textbf{4.00}} & \revnum{3.63} & \revnum{\textbf{3.38}} & \revnum{3.75} & \revnum{3.59} & \revnum{\textbf{3.13}} & \revnum{3.50} & \revnum{3.13} & \revnum{3.03} & \revnum{\textbf{3.56}}\\
\midrule
PAVG & \revnum{3.72} & \revnum{3.72} & \revnum{3.97} & \revnum{\textbf{3.66}} & \revnum{3.28} & \revnum{\textbf{3.81}} & \revnum{\textbf{3.63}} & \revnum{3.09} & \revnum{3.50} & \revnum{\textbf{3.22}} & \revnum{\textbf{3.22}} & \revnum{\textbf{3.56}}\\
\bottomrule
\end{tabular*}
\end{table}

\paragraph{Results by scene type.}
\revnew{Scores averaged over the same four judges are reported separately for eight solid, eight gas, and sixteen solid--gas scenes in Table~\ref{tab:video-by-phase}. Each scene has equal weight within its subset. The overall scores in main-paper Table~1 therefore weight these subsets by $1/4$, $1/4$, and $1/2$, respectively.}

\begin{table}[htbp]
\centering
\caption{Video scores by scene type, averaged over four judges.
\textbf{Bold}: best within each subset and metric.}
\label{tab:video-by-phase}
\small
\setlength{\tabcolsep}{3.2pt}
\renewcommand{\arraystretch}{1.08}
\begin{tabular*}{\linewidth}{@{\extracolsep{\fill}}l*{9}{c}@{}}
\toprule
 & \multicolumn{3}{c}{Solid (8 scenes)}
 & \multicolumn{3}{c}{Gas (8 scenes)}
 & \multicolumn{3}{c}{Solid--gas (16 scenes)}\\
\cmidrule(lr){2-4}\cmidrule(lr){5-7}\cmidrule(lr){8-10}
Method & SA$\uparrow$ & PC$\uparrow$ & VQ$\uparrow$
 & SA$\uparrow$ & PC$\uparrow$ & VQ$\uparrow$
 & SA$\uparrow$ & PC$\uparrow$ & VQ$\uparrow$\\
\midrule
HunyuanVideo-1.5 & \revnum{3.75} & \revnum{3.50} & \revnum{3.72} & \revnum{2.44} & \revnum{2.66} & \revnum{3.41} & \revnum{2.98} & \revnum{3.05} & \revnum{\textbf{3.77}}\\
Wan2.2-I2V-A14B & \revnum{3.47} & \revnum{3.09} & \revnum{3.53} & \revnum{2.84} & \revnum{3.19} & \revnum{\textbf{3.69}} & \revnum{2.52} & \revnum{2.72} & \revnum{3.47}\\
CogVideoX1.5-5B-I2V & \revnum{1.81} & \revnum{1.78} & \revnum{2.34} & \revnum{2.59} & \revnum{3.09} & \revnum{3.56} & \revnum{2.14} & \revnum{2.25} & \revnum{2.94}\\
DragAnything & \revnum{1.13} & \revnum{1.31} & \revnum{2.19} & \revnum{3.09} & \revnum{\textbf{3.34}} & \revnum{3.22} & \revnum{2.44} & \revnum{2.48} & \revnum{2.73}\\
ObjCtrl-2.5D & \revnum{1.91} & \revnum{1.94} & \revnum{2.50} & \revnum{2.66} & \revnum{2.75} & \revnum{2.91} & \revnum{2.28} & \revnum{2.36} & \revnum{2.91}\\
FlashMotion & \revnum{3.19} & \revnum{3.09} & \revnum{3.63} & \revnum{2.66} & \revnum{3.19} & \revnum{3.38} & \revnum{3.20} & \revnum{3.17} & \revnum{3.63}\\
Wan-Move & \revnum{2.72} & \revnum{2.72} & \revnum{2.84} & \revnum{1.44} & \revnum{1.34} & \revnum{1.72} & \revnum{2.05} & \revnum{1.89} & \revnum{2.25}\\
Force Prompting & \revnum{1.19} & \revnum{1.34} & \revnum{2.31} & \revnum{2.22} & \revnum{2.78} & \revnum{2.84} & \revnum{2.34} & \revnum{2.36} & \revnum{3.05}\\
StreamForce & \revnum{1.22} & \revnum{1.34} & \revnum{2.38} & \revnum{2.47} & \revnum{2.81} & \revnum{2.97} & \revnum{1.92} & \revnum{2.02} & \revnum{2.77}\\
PhysCtrl & \revnum{\textbf{3.78}} & \revnum{\textbf{3.56}} & \revnum{\textbf{3.91}} & \revnum{3.38} & \revnum{3.31} & \revnum{3.66} & \revnum{3.47} & \revnum{3.22} & \revnum{3.63}\\
\midrule
PAVG & \revnum{3.59} & \revnum{3.41} & \revnum{3.78} & \revnum{\textbf{3.44}} & \revnum{\textbf{3.34}} & \revnum{3.63} & \revnum{\textbf{3.59}} & \revnum{\textbf{3.28}} & \revnum{3.72}\\
\bottomrule
\end{tabular*}
\end{table}
\FloatBarrier

\paragraph{Shared scoring rubric.}
The frozen rubric retains PhysCtrl's three criteria and five-point scale, with explicit guidance for gas continuity, torque, and temporal evidence.
The following template summarizes the request shared by all four judges; the scene-specific text prompt is appended after it.
\begin{quote}
\small
Evaluate 11 anonymous image-to-video results for the same scene.
Attachment 1 is the annotated input image; the corresponding clean image was used for generation.
The arrows are not objects to be generated.
Attachments 2--5 represent Video 1, 6--9 represent Video 2, and so on through 42--45 for Video 11.
Read each video's four sheets in order, with samples 1--3, 4--6, 7--8, and 9--10 arranged top to bottom.
Each tile is a full frame; its position on the sheet is not motion.

Score each criterion independently on a five-point Likert scale (1 = poor, 5 = excellent).
Inspect all ten samples across all four sheets before scoring.
Track displacement, orientation, support contact, and gas against stable landmarks through early, middle, and late samples, accounting for camera motion.
Small, gradual, or out-and-back motion is not no motion; do not judge from one frame or only the endpoints.
If motion is unclear at the supplied resolution, state uncertainty rather than claiming that nothing moves.
\begin{itemize}
\item \textbf{Semantic Adherence (SA):} How well content and motion match the text prompt, especially the force direction and position.
The video should start from the input image.
For gas, check the specified motion and interaction and trace later clouds to initial gas or a scene/prompt-supported source.
Substantial new clouds or abrupt replacement without a plausible origin are errors even when their direction is correct.
Normal deformation, diffusion, splitting, and emergence from occlusion need not preserve the initial silhouette.
\item \textbf{Physical Commonsense (PC):} Whether motion is physically plausible given the force direction and application point.
Consider translation and torque about the center of mass or relevant support/pivot.
Penalize clearly missing or opposite rotation when expected and visible, but not plausible torque-driven tilt.
Do not require rotation when gravity, contact, constraints, or a negligible moment arm explain its absence; state uncertainty for ambiguous geometry.
For gas, assess plausible transport, deformation, spreading, and solid--gas interaction from the initial conditions.
Penalize unsupported gas creation, solid-like behavior, and sticker-like translation that preserves the gas shape and internal pattern throughout without plausible evolution or interaction.
Bulk advection alone is not an error, and dramatic turbulence is not required.
\item \textbf{Video Quality (VQ):} Overall visual and temporal quality; static or nearly-static sequences are less preferred.
Assess visible artifacts and discontinuities independently: smooth rendering does not excuse semantic or physical errors.
\end{itemize}
Return a JSON object containing a \texttt{videos} array with exactly eleven entries.
Each contains \texttt{video}, a concise Chinese \texttt{observed\_motion} summary of early-to-middle-to-late changes, and \texttt{SA}, \texttt{PC}, and \texttt{VQ} objects.
Each criterion contains a brief Chinese \texttt{reason}, one to four supporting sample numbers in \texttt{samples}, and an integer \texttt{score} from 1 to 5.
Support motion or stillness claims with early and late samples, checking intermediate samples too.
\end{quote}
No prompt specifies a preferred method, a required ranking, or a target score distribution.

\paragraph{Interpretation and limitations.}
These ratings measure perceived adherence, plausibility, and quality, not physical ground truth.
Finite temporal sampling and image resizing can obscure subtle motion or artifacts.
Uniform sampling compares visible progression; it does not establish equal physical speeds across backbones with different native temporal settings.
The scores characterize this frozen video release and rubric, and should not be treated as directly comparable to ratings from earlier releases.

\clearpage
\section{\revnew{Additional Evaluation Results}}
\label{app:additional-results}

\subsection{\revnew{Human Preference Study}}
\label{app:human-preference}
We conducted a user study to evaluate the physics plausibility and overall quality of videos generated by PAVG and ten baselines.
\revnew{This study uses the v8 video release with the current PAVG and updated PhysCtrl clips; the remaining baseline clips are retained from the frozen comparison release.}
Each participant evaluated 32 scenes, each containing an annotated input image, a text prompt describing the scene and physical conditions, and eleven generated results presented as two-second looping GIFs.
Red arrows indicate forces on solids, while blue arrows indicate gas initial velocities.
Arrow tails mark the force application point or initial-velocity annotation location, and arrowheads indicate the corresponding direction.
Scene and method orders were randomized, with method identities hidden.

\begin{figure}[H]
\centering
\includegraphics[width=0.55\linewidth]{\pavgroot/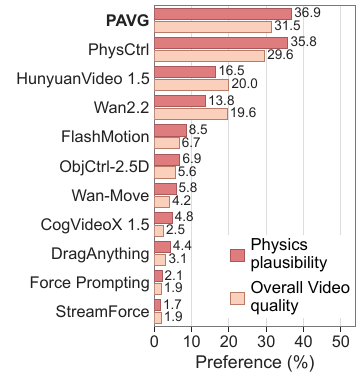}
\caption{Human preference rates \revnew{on the v8 video release}, averaged over 32 scenes for eleven methods (15 questionnaires; 480 participant--scene responses per criterion).
Participants can select multiple equally best videos per criterion, so rates across methods can sum to over 100\%.}
\label{fig:human-preference}
\end{figure}

Participants were asked to carefully observe the results and select one or more equally best videos independently for two criteria:
(1) \emph{Physics plausibility}: motion that best matches the text prompt and solid-force or gas-initial-velocity direction, considering both motions and their interaction when both phases are present.
The arrows and text prompts were assumed to match.
(2) \emph{Overall Video quality}: the best visual and temporal quality.

We received 15 complete questionnaires, yielding $15 \times 32 = 480$ scene evaluations per criterion.
For each question, we computed each method's preference rate as its number of selections divided by 480, expressed as a percentage.
Each selected method was counted once per scene, including tied choices.
\revnew{As shown in Fig.~\ref{fig:human-preference}, PAVG ranks first among the 11 methods in both physical plausibility (36.9\% preference) and overall video quality (31.5\%).}

\subsubsection*{Particle dynamics in the 32 video scenes}
To complement the video ratings, we compare the particle trajectories predicted by PAVG and PhysCtrl against a common simulation of each scene.

\paragraph{Simulation reference.}
Each simulation starts from the same reconstructed points and prescribed controls as the two predictors. We use material point method (MPM) rollouts for deformable solids and a MuJoCo convex hull for the rigid object, retaining the force mask and contact plane where enabled. Seven interaction scenes use stiff elastic solids whose original rigid-body inputs stored zero particle volumes. For their MPM references, we divide each occupied grid cell's volume equally among its particles and use the resulting total mass and number of forced particles to scale the prescribed force vector. Gas tracers follow a PhiFlow semi-Lagrangian velocity field with pressure projection on a padded periodic grid. The prescribed velocity tapers to still ambient air outside the initial cloud. In interaction scenes, the nearest simulated solid sample transfers momentum locally to the gas. These references describe the reconstructed geometry and assumed material and flow conditions of each image.

\paragraph{What the metrics measure.}
We use kinetic-energy MSE from \href{https://proceedings.neurips.cc/paper_files/paper/2023/hash/ccac3b120c7dc86d45f56830732b62be-Abstract-Datasets_and_Benchmarks.html}{LagrangeBench (Toshev et al., NeurIPS 2023)} to measure motion intensity. For each phase, we average squared speed over 2,048 equal-weight particles in model units, then compute the mean squared difference between the predicted and reference energy curves. Reversing every velocity leaves this energy unchanged.
To measure direction, we compare the predicted and simulated displacement of each particle over each interval. We average their direction cosines, weighting by simulated displacement length. A score of $1$ indicates alignment, $0$ orthogonality for an individual pair, and $-1$ opposite directions. A stationary reference step has zero weight and a stationary prediction contributes zero to the numerator.

We use the 24 forecast states at stored indices $1,3,\ldots,47$ and the simulation clock (0.05 per stored interval for gas-only scenes and 0.0417 otherwise). LagrangeBench computes energy over the 23 intervals between forecast states. Direction includes the first displacement from the initial state. Each scene has equal weight, with both phases weighted equally in interaction averages.

\begin{table}[H]
\centering
\caption{Kinetic-energy MSE (Energy, $\times 100$) and direction cosine (Cosine) against the simulation reference ($40^3$ gas grid).}
\label{tab:video32-physics}
\small
\setlength{\tabcolsep}{6pt}
\begin{tabular}{@{}lcccccc@{}}
\toprule
Method & \multicolumn{2}{c}{Solid (8)} & \multicolumn{2}{c}{Gas (8)} & \multicolumn{2}{c}{Interaction overall (16)} \\
\cmidrule(lr){2-3}\cmidrule(lr){4-5}\cmidrule(lr){6-7}
& Energy$\downarrow$ & Cosine$\uparrow$ & Energy$\downarrow$ & Cosine$\uparrow$ & Energy$\downarrow$ & Cosine$\uparrow$ \\
\midrule
PhysCtrl & \textbf{0.31} & 0.893 & 4.59 & \textbf{0.630} & 5.94 & 0.546 \\
PAVG & 0.79 & \textbf{0.924} & \textbf{4.27} & 0.537 & \textbf{2.89} & \textbf{0.710} \\
\bottomrule
\end{tabular}
\end{table}

The methods are competitive on single-phase scenes (Table~\ref{tab:video32-physics}). On solids, PAVG predicts motion direction more accurately, while PhysCtrl has lower kinetic-energy error. On gas, PAVG has lower energy error and PhysCtrl better matches the flow direction. The clearest advantage for PAVG appears in solid--gas interactions, as in the main trajectory benchmark (main-paper Table~\ref{tab:overall-trajectory}). Across the 16 interactions, PAVG reduces mean kinetic-energy error by 51.4\% and increases mean direction cosine by 30.1\% relative to PhysCtrl. Its predicted interactions thus more closely match both the intensity and direction of the simulated motion.

\FloatBarrier
\subsection{\revnew{Liquid Extension}}
\label{app:liquid-extension}

\paragraph{\revnew{Setting and training.}}
\revnew{We retain the original single-branch trajectory backbone: eight Transformer blocks, width~256, four heads, 2,048 input points, and 24 future states.
The static gravity/force interface is retained, while material conditioning uses liquid density and dynamic viscosity.
Parameter-free floor-relative coordinates and displacement scales fitted on training data retain the backbone's parameter count.
The preliminary model is trained separately for 60,000 updates on 11,167 gravity-only liquid trajectories, with 635 validation examples; the 663-example test split is excluded.
The four materials are water, milk, edible oil, and honey.
The examples involve gravity-driven floor impact and spreading, with additional force set to zero.
Training uses axis-balanced position and interval-velocity losses, a floor penalty, and a GT-contact height term weighted by~10.
This experiment requires liquid training and the stated supervision; it is not zero-shot transfer of the gas checkpoint.}

\begin{figure}[H]
\centering
\includegraphics[width=\linewidth]{\pavgroot/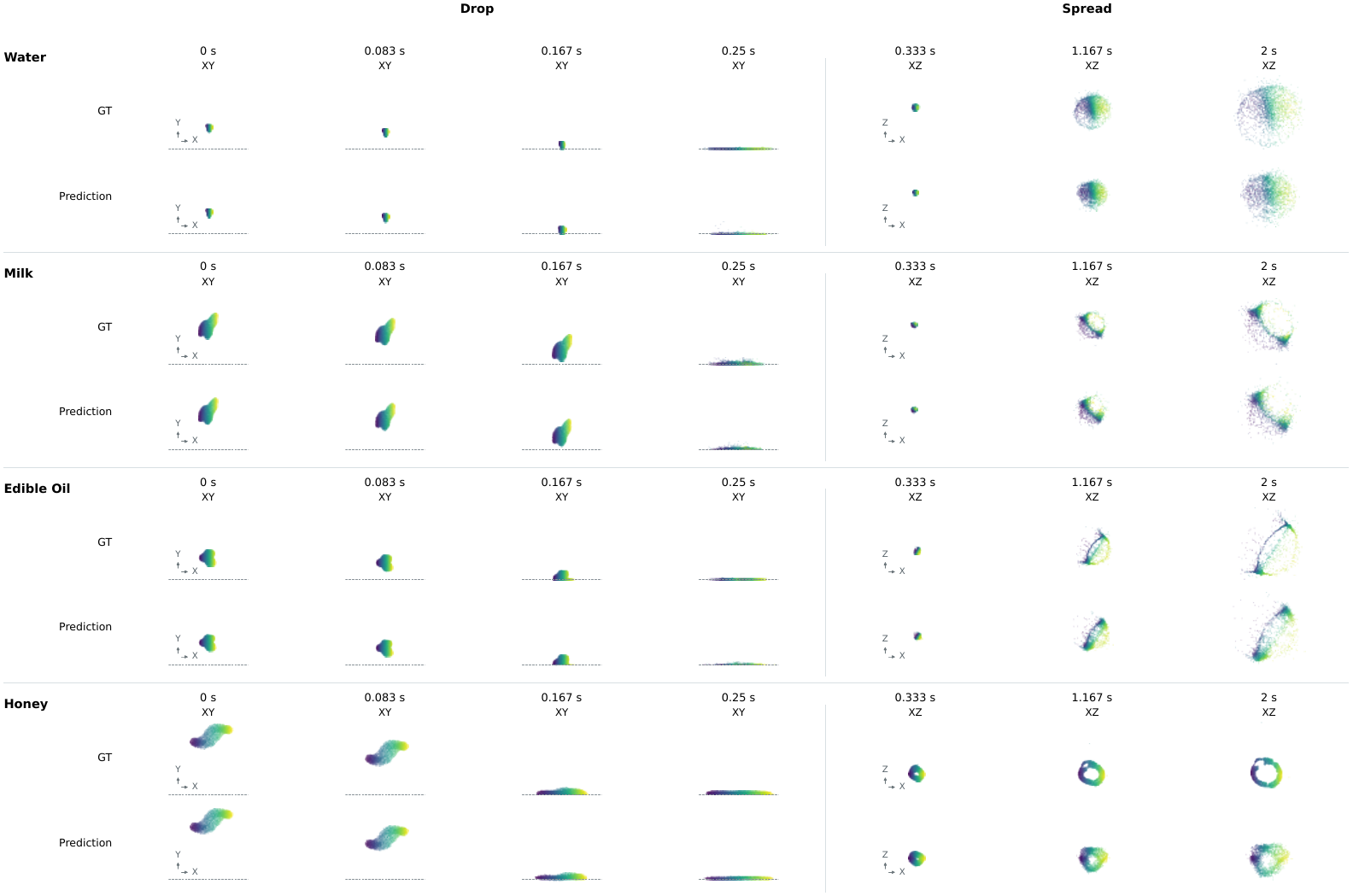}
\caption{\revnew{Preliminary liquid-trajectory extension on four validation examples. Each material separates the fast drop phase (0, 0.083, 0.167, and 0.25~s), shown in XY side views, from the subsequent spreading phase (0.333, 1.167, and 2~s), shown in XZ top views. GT is above prediction. Y is height above the known floor (dashed line), and equal physical axis scales are used within each phase. Colors track initial point identities. All 25 saved states are shown in the supplementary media (Section~3).}}
\label{fig:discussion-liquid-extension}
\end{figure}

\paragraph{\revnew{Qualitative results.}}
\revnew{Four saved validation examples are compared with GT over a two-second horizon in Fig.~\ref{fig:discussion-liquid-extension}.
The figure samples the fast drop phase and the later spreading phase separately, using uniform times within each phase.
XY projections show vertical motion during the drop, while XZ projections show horizontal spreading after contact, using the same point identities and initial-state colors throughout.
The coordinate normalization is inverted and the known floor defines zero height on the Y axis; equal physical axis scales and fixed limits are used across GT, prediction, and time within each material.
Predictions are not projected onto the floor or aligned to GT.
The predictor captures the downward settling and horizontal spreading of the liquid point clouds, with material-dependent spatial patterns across water, milk, edible oil, and honey.
Local dispersion differs from GT, particularly for honey, while the shared backbone reproduces the overall motion trend after liquid-specific training.
These qualitative results support extending PAVG's trajectory modeling to liquids.
Full GT/prediction GIFs and the corresponding numerical arrays are available in the supplementary media (Section~3).}

\FloatBarrier
\subsection{\revnew{Model Capacity and Inference Cost}}
\label{app:capacity-inference}

\revnew{We compare PAVG with two shared-backbone trajectory predictors using the same training data and source exposure budgets (Sec.~\ref{app:ablation-settings}). Retrained PhysCtrl matches PAVG's single-phase inference cost to within 0.4\%. To match PAVG's total parameter count, we expand the unified phase predictor from eight layers of width 256 to nine layers of width 384. The resulting model has 0.174\% fewer parameters than PAVG (Table~\ref{tab:app-capacity-cost}).}

\begin{table}[H]
\centering
\caption{Trajectory predictor configurations and inference cost per scene and denoising step.}
\label{tab:app-capacity-cost}
\small
\setlength{\tabcolsep}{3pt}
\begin{tabular*}{\linewidth}{@{\extracolsep{\fill}}llrrr r rrr@{}}
\toprule
& & \multicolumn{3}{c}{Per backbone} & & \multicolumn{3}{c}{GFLOPs per step} \\
\cmidrule(lr){3-5}\cmidrule(l){7-9}
Predictor & Backbone & Layers & Width & Heads & Parameters & Solid only & Gas only & Interaction \\
\midrule
PAVG & Two, with CPI & 8 & 256 & 4 & 40,641,830 & 1,743.855 & 1,740.394 & 6,082.201 \\
Unified phase predictor & Shared & 9 & 384 & 6 & 40,571,139 & 3,676.691 & 3,676.691 & 7,353.382 \\
PhysCtrl (retrained) & Shared & 8 & 256 & 4 & 16,096,003 & 1,746.476 & 1,746.476 & 3,487.721 \\
\bottomrule
\end{tabular*}
\end{table}

\paragraph{\revnew{Inference and phase weighting.}}
\revnew{Each predictor takes 2,048 points per present phase and predicts 24 future frames using 25 DDIM steps. PAVG runs only the corresponding Stage~2 branch for a single phase and uses both branches with bidirectional CPI for interactions. Each shared model runs its backbone once per present phase.}
\revnew{After splitting interactions, the shared backbone sees 3,569,294 solid examples and 2,461,340 gas examples over training. With equal per-example weights, solid motion would contribute more terms to the shared objective. We weight the loss inversely to the number of exposures for each phase (0.844794 for solids and 1.225071 for gas), giving the two phases equal total sample weight. PAVG pretrains a separate branch for each phase. At Stage~2, single-phase examples supervise the corresponding branch and paired examples train both branches through CPI. PAVG uses unit weights for the two phase losses.}

\begin{table}[H]
\centering
\caption{Trajectory accuracy on 700 held-out scenes. Interaction metrics are computed on the combined solid and gas point clouds.}
\label{tab:app-capacity-accuracy}
\small
\setlength{\tabcolsep}{2pt}
\begin{tabular*}{\linewidth}{@{\extracolsep{\fill}}l rrr rrr rrr@{}}
\toprule
& \multicolumn{3}{c}{Solid only (500)} & \multicolumn{3}{c}{Gas only (100)} & \multicolumn{3}{c}{Interaction overall (100)} \\
\cmidrule(lr){2-4}\cmidrule(lr){5-7}\cmidrule(l){8-10}
Predictor & vIoU $\uparrow$ & CD $\downarrow$ & Corr $\downarrow$ & vIoU $\uparrow$ & CD $\downarrow$ & Corr $\downarrow$ & vIoU $\uparrow$ & CD $\downarrow$ & Corr $\downarrow$ \\
\midrule
PAVG & 0.627285 & 0.018528 & 0.066111 & 0.793380 & 0.001353 & 0.061098 & \textbf{0.700409} & \textbf{0.003384} & \textbf{0.056061} \\
Unified phase predictor & \textbf{0.632757} & \textbf{0.016061} & \textbf{0.061839} & \textbf{0.794432} & \textbf{0.001314} & \textbf{0.058881} & 0.567582 & 0.023202 & 0.171732 \\
PhysCtrl (retrained) & 0.613280 & 0.022733 & 0.071951 & 0.783158 & 0.001423 & 0.065287 & 0.552442 & 0.023672 & 0.180476 \\
\bottomrule
\end{tabular*}
\end{table}

\paragraph{\revnew{Single-phase and interaction accuracy.}}
\revnew{At comparable single-phase inference cost, PAVG achieves higher vIoU than retrained PhysCtrl on all three subsets (Table~\ref{tab:app-capacity-accuracy}). The capacity-matched unified predictor uses 2.11 times as many FLOPs as PAVG on both single-phase subsets, with vIoU gains of 0.0055 on solids and 0.0011 on gas. On interactions, PAVG reaches 0.7004 vIoU compared with 0.5676 for the unified predictor and also yields lower Chamfer distance and correspondence error. Increasing shared capacity therefore brings single-phase vIoU close to PAVG while leaving a larger gap on interacting trajectories.}

\end{document}